%% file: main.tex
\documentclass[10pt,twocolumn,letterpaper]{article}
\usepackage[letterpaper,margin=0.72in,columnsep=0.28in]{geometry}
\usepackage{fontspec}
\usepackage{unicode-math}
\usepackage{amsmath}
\usepackage{newunicodechar}
\newunicodechar{∈}{\ensuremath{\in}}\newunicodechar{∖}{\ensuremath{\setminus}}
\newunicodechar{∪}{\ensuremath{\cup}}\newunicodechar{⊆}{\ensuremath{\subseteq}}
\newunicodechar{ℝ}{\ensuremath{\mathbb{R}}}\newunicodechar{∅}{\ensuremath{\emptyset}}
\newunicodechar{×}{\ensuremath{\times}}\newunicodechar{≤}{\ensuremath{\le}}
\newunicodechar{≥}{\ensuremath{\ge}}\newunicodechar{≈}{\ensuremath{\approx}}
\newunicodechar{≠}{\ensuremath{\neq}}\newunicodechar{∥}{\ensuremath{\|}}
\newunicodechar{ρ}{\ensuremath{\rho}}\newunicodechar{τ}{\ensuremath{\tau}}
\newunicodechar{λ}{\ensuremath{\lambda}}\newunicodechar{ξ}{\ensuremath{\xi}}
\newunicodechar{ψ}{\ensuremath{\psi}}\newunicodechar{θ}{\ensuremath{\theta}}
\newunicodechar{Δ}{\ensuremath{\Delta}}\newunicodechar{σ}{\ensuremath{\sigma}}
\newunicodechar{⁻}{\textsuperscript{$-$}}\newunicodechar{⁰}{\textsuperscript{0}}
\newunicodechar{⁴}{\textsuperscript{4}}\newunicodechar{⁵}{\textsuperscript{5}}
\newunicodechar{⁶}{\textsuperscript{6}}\newunicodechar{⁸}{\textsuperscript{8}}
\newunicodechar{₀}{\textsubscript{0}}\newunicodechar{₂}{\textsubscript{2}}
\newunicodechar{✓}{\ensuremath{\checkmark}}\newunicodechar{✗}{\ensuremath{\times}}
\usepackage{calc}\usepackage{array}\usepackage{longtable}\usepackage{booktabs}
\usepackage{graphicx}\usepackage{rotating}\usepackage{caption}
\usepackage{xcolor}
\definecolor{accent}{RGB}{38,78,120}\definecolor{linkblue}{RGB}{28,94,170}
\usepackage[colorlinks=true,urlcolor=linkblue,linkcolor=accent,citecolor=accent]{hyperref}
\usepackage{microtype}\usepackage{titlesec}\usepackage{fancyhdr}
\titleformat{\section}{\normalfont\large\bfseries\color{accent}}{}{0pt}{\raisebox{-0.03em}{\textcolor{accent}{\rule{0.62em}{0.62em}}}\hspace{0.55em}}
\titleformat{\subsection}{\normalfont\normalsize\bfseries\color{accent!85!black}}{}{0pt}{}
\titlespacing*{\section}{0pt}{2.6ex plus .4ex}{1.0ex}
\titlespacing*{\subsection}{0pt}{1.8ex plus .3ex}{0.6ex}
\fancypagestyle{plain}{\fancyhf{}}

\begin{document}
\twocolumn[%
  \vspace*{1.2em}
  \begin{center}
  {\Large\bfseries M-JEPA: Predictive Self-Supervised Learning for Molecular
  Graphs with Scaffold-Shift Evaluation on Tox21}\\[8pt]
  {\large Karthik Iyer\quad Nasser Sabar}\\[4pt]
  {\small Department of Computer Science and Information Technology,
  La Trobe University, Melbourne, VIC 3086, Australia}\\[2pt]
  {\footnotesize Corresponding author: \texttt{karthik.iyer.au@gmail.com}}
  \end{center}\vspace{4pt}
  \begin{center}\begin{minipage}{0.86\textwidth}
  {\small\textbf{Abstract}\\[3pt]\input{abstract}}
  \vspace{6pt}\\
  {\footnotesize\itshape This document is the unedited author's version of a
  Submitted Work that was subsequently accepted for publication in the
  \upshape Journal of Chemical Information and Modeling\itshape,
  \textcopyright{} American Chemical Society, after peer review. To access the
  final edited and published work, see
  \upshape\href{https://doi.org/10.1021/acs.jcim.6c00828}{doi.org/10.1021/acs.jcim.6c00828}\itshape.}
  \end{minipage}\end{center}\vspace{16pt}
]
\input{body}
\input{si}
\end{document}

%% file: abstract.tex
Self-supervised molecular representation learning is increasingly used
to improve transfer on small, label-limited property prediction tasks,
yet commonly used contrastive objectives can be sensitive to view
construction and are often evaluated with protocols that report only
discrimination. We adapt Joint Embedding Predictive Architectures
(JEPAs) to molecular graphs by masking chemically coherent connected
subgraphs and training a context encoder to regress the masked target
embedding produced by an exponential-moving-average (EMA) teacher. To
evaluate predictive self-supervision under realistic distribution shift,
we introduce a leakage-resistant, compute-matched three-phase protocol:
(i) an 80-run objective screening on ESOL with fixed compute budgets,
(ii) JEPA-only refinement of masking and augmentation choices, and (iii)
per-assay Tox21 evaluation under Bemis--Murcko scaffold splits with
joint discrimination--reliability analysis and attribution diagnostics.
Under matched compute, M-JEPA achieves substantially lower ESOL proxy
RMSE than a compute-matched InfoNCE baseline (mean RMSE 2.12 vs 3.75;
mean paired ΔRMSE 1.61; 95\% bootstrap CI {[}1.52, 1.72{]}; Wilcoxon
p=2.4×10\^{}−4) and typically completes runs faster (≈15--25 min vs
≈35--45 min on a single GPU under identical hardware), albeit with
higher runtime variance. On Tox21, improvements in ROC-AUC/PR-AUC are
assay-dependent and do not uniformly imply better calibration;
reliability diagrams and Brier/ECE metrics reveal endpoints where
fine-tuning improves ranking but leaves probabilities miscalibrated.


%% file: body.tex
\section{Introduction}

Accurate prediction of chemical properties is central to cheminformatics
workflows, from prioritizing compounds in virtual screening to assessing
safety liabilities early in discovery. Supervised models can perform
well when labels are abundant and evaluation distributions match
deployment, but many practical tasks are label-limited and experience
strong distribution shift when novel scaffolds are encountered.

Graph neural networks (GNNs) have become a standard backbone for
molecular property prediction because they operate directly on
atom--bond graphs and can learn task-relevant representations via
message passing (e.g., D-MPNN, GINE, and related architectures).
However, training such models from scratch on small endpoints is prone
to overfitting and can yield unreliable probability estimates under
shift. (Kearnes et al., 2016; Gilmer et al., 2017; Withnall et al.,
2020; Xiong et al., 2020).

Self-supervised learning (SSL) addresses data scarcity by pretraining
representations on large unlabeled molecule corpora before fine-tuning.
Many influential molecular SSL approaches are contrastive, using
augmented `views' of the same molecule and an InfoNCE objective to
encourage invariance while separating negatives (e.g., MolCLR and
related graph contrastive formulations). In chemistry, however,
aggressive augmentations can violate chemical `grammar' (e.g., by
perturbing functional groups or stereochemistry) and may erase
predictive signal rather than improve invariance. (Wang et al., 2021; Li
et al., 2021; Hu et al., 2020; Wu et al., 2023).

Related work and positioning. Molecular SSL has been explored via
contrastive objectives (e.g., MolCLR, GeomGCL), generative or masked
modeling (e.g., self-supervised graph transformers; Mol-BERT), and
predictive objectives that regress representations rather than
discriminate negatives. Recent surveys categorize graph SSL into
contrastive, generative, and predictive families. M-JEPA is positioned
in the predictive family and emphasizes chemically coherent
connected-subgraph masking together with leakage-resistant,
compute-matched evaluation that jointly reports discrimination,
calibration, and attribution diagnostics under scaffold shift. (Wang et
al., 2021; Li et al., 2021; Rong et al., 2020; Li and Jiang, 2021; Wu et
al., 2023; Skenderi et al., 2024; Piccoli et al., 2025).

Joint Embedding Predictive Architectures (JEPAs) provide a
non-contrastive alternative: a student predicts a latent representation
of a masked target region produced by a momentum-updated teacher,
replacing explicit negative sampling with embedding-space regression.
This predictive formulation is appealing for molecules because
chemically coherent masking can create a structured prediction
problem---recovering a missing substructure from its molecular
context---while avoiding the need to define many augmentation-driven
positives and negatives. (Assran et al., 2023; Skenderi et al., 2024;
Piccoli et al., 2025).

In this work we propose M-JEPA, a JEPA adaptation for molecular graphs
with connected-subgraph masking and an EMA teacher, and we evaluate it
with a leakage-resistant pipeline designed to reduce selection bias. Our
contributions are: (1) a molecular JEPA objective with chemically
coherent masking and compute-matched contrastive baselines; (2) a
three-phase evaluation protocol separating objective screening,
configuration selection, and downstream testing under Bemis--Murcko
scaffold splits; (3) reliability-aware reporting on Tox21 using
ROC-AUC/PR-AUC together with Brier score, expected calibration error
(ECE), and reliability diagrams; and (4) motif-level integrated-gradient
attribution diagnostics to support qualitative interpretation under
scaffold shift. Code and experiment artifacts are available at
https://github.com/Kar488/M-JEPA/ (currently private; available to
reviewers upon request; will be made public on publication). (Bemis and
Murcko, 1996; Wu et al., 2017; Sundararajan et al., 2017; Guo et al.,
2017; Brier, 1950).

\section{Methods}

We pretrain graph encoders on a 10M-molecule subset of ZINC and evaluate
transfer in a controlled three-phase protocol. Phase 1 performs an
80-run, compute-capped objective screening on ESOL validation RMSE to
compare JEPA against a compute-matched contrastive baseline without
tuning on downstream labels. Phase 2 refines masking and augmentation
choices within the JEPA family under the same compute constraints. Phase
3 evaluates the selected configuration on Tox21 using Bemis--Murcko
scaffold splits and per-assay reporting of discrimination and
calibration. Downstream (Tox21) labels and metrics are not accessed
during Phases 1--2; objective-family and configuration ranking are
performed exclusively using the ESOL validation proxy, and Phase 3 is
executed once with the locked selection. (Sterling and Irwin, 2015;
Delaney, 2004; Wu et al., 2017; Bemis and Murcko, 1996).

M-JEPA uses a student (context) encoder, a momentum-updated EMA teacher
(target) encoder, and a predictor MLP. For each molecule graph G, we
sample a connected node set S by breadth-first expansion from a random
seed until a target mask ratio is reached. Connected masking preserves
chemically coherent fragments (e.g., ring systems and functional groups)
and yields a structured missing-substructure prediction task. In
contrast, random atom masking can produce disconnected targets that are
less chemically interpretable and may be easier for message-passing
models to infer from local neighborhoods. We then form two related
views: a context view C = G{[}V \textbackslash{} S{]} and a target view
T = G{[}S{]} (induced subgraph). The student embeds C, the teacher
embeds T (no gradients), and the predictor regresses teacher embeddings
from student embeddings using an embedding-space regression loss with
light regularization. (Assran et al., 2023; Skenderi et al., 2024).

We apply view perturbations through predefined augmentation profiles
(e.g., none, graph\_light, geom\_only, mixed\_light). Augmentations are
applied independently to the student and teacher branches in JEPA; the
contrastive baseline uses the same view family to form positive pairs
under an InfoNCE objective with a projection head. All baselines share
backbones (D-MPNN, GINE, SchNet3D) and are compute-matched by fixing
pretraining budget (2 epochs, max\_pretrain\_batches = 800) and
downstream fine-tuning budget. (Wang et al., 2021; Li et al., 2021; Yang
et al., 2019; Schutt et al., 2017).

Downstream evaluation follows standard practice for binary endpoints
with strong class imbalance: we report ROC-AUC and PR-AUC for
discrimination, and we quantify reliability with the Brier score and
expected calibration error (ECE) computed on predicted probabilities.
Reliability diagrams and per-assay calibration curves are used to
diagnose mismatch between ranking and probability quality under scaffold
shift. (Brier, 1950; Guo et al., 2017).

For interpretability, we compute integrated gradients (IG) on fine-tuned
models and aggregate atom-level attributions to chemically meaningful
motifs derived from ring systems and Bemis--Murcko scaffold components.
We report representative examples selected by a fixed protocol
(Supporting Information) and include stability diagnostics (e.g.,
attribution rank agreement across checkpoints) to reduce subjectivity in
qualitative interpretation. (Sundararajan et al., 2017;
Sanchez-Lengeling et al., 2020; Ying et al., 2019; Simpson et al.,
2025).

\begin{figure*}[tp]
\centering
\includegraphics[width=0.92\textwidth]{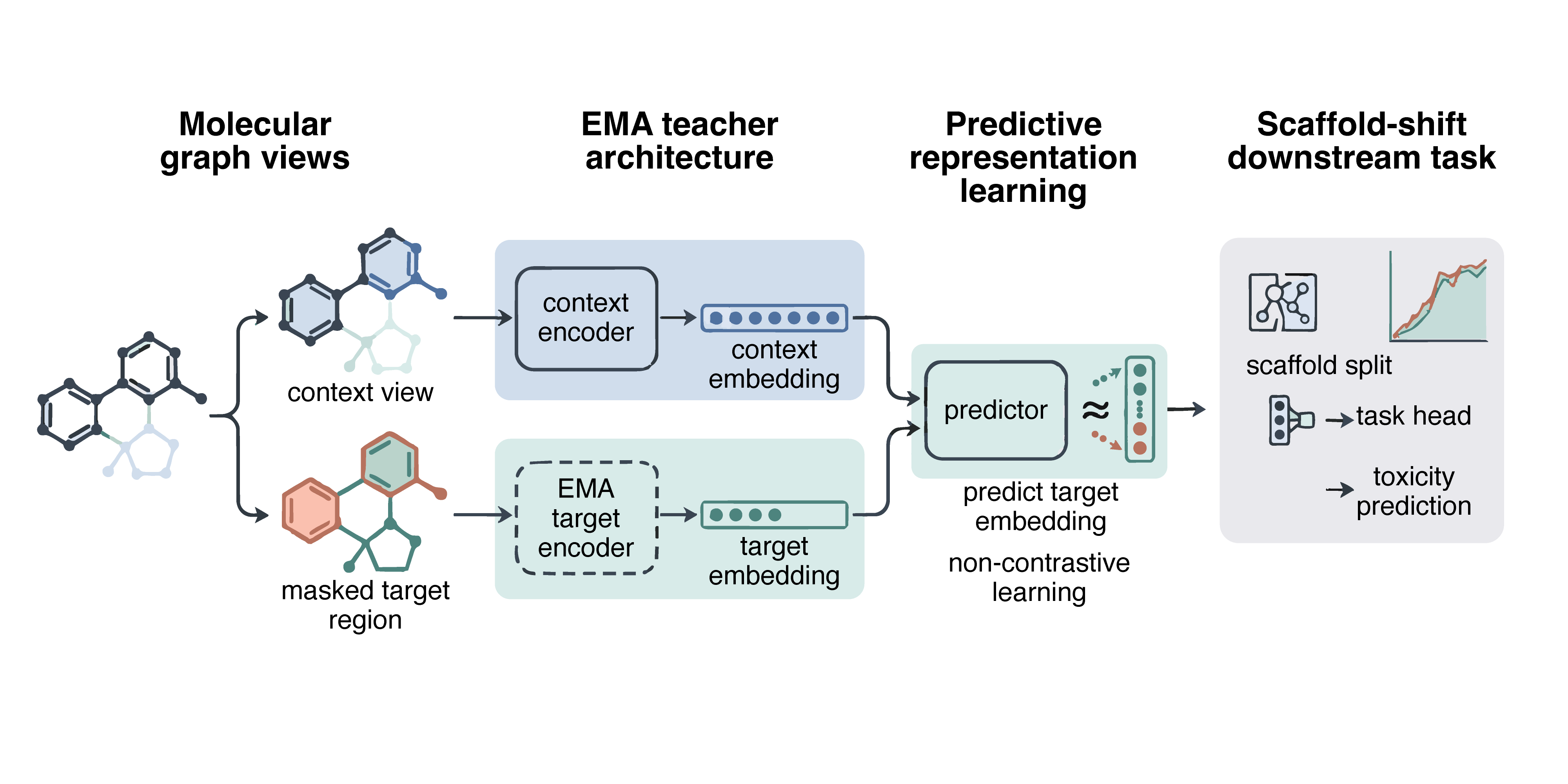}
\caption{Overview of predictive self-supervised learning on
molecular graphs. A chemically coherent substructure is masked from a
molecular graph, and a context encoder is trained to predict the latent
representation of the masked target region produced by an
exponential-moving-average (EMA) teacher. The resulting representations
support downstream molecular property prediction with subgraph-level
attribution diagnostics and calibrated probability estimates under
scaffold-split evaluation.}
\end{figure*}

\begin{figure*}[tp]
\centering
\includegraphics[width=0.92\textwidth]{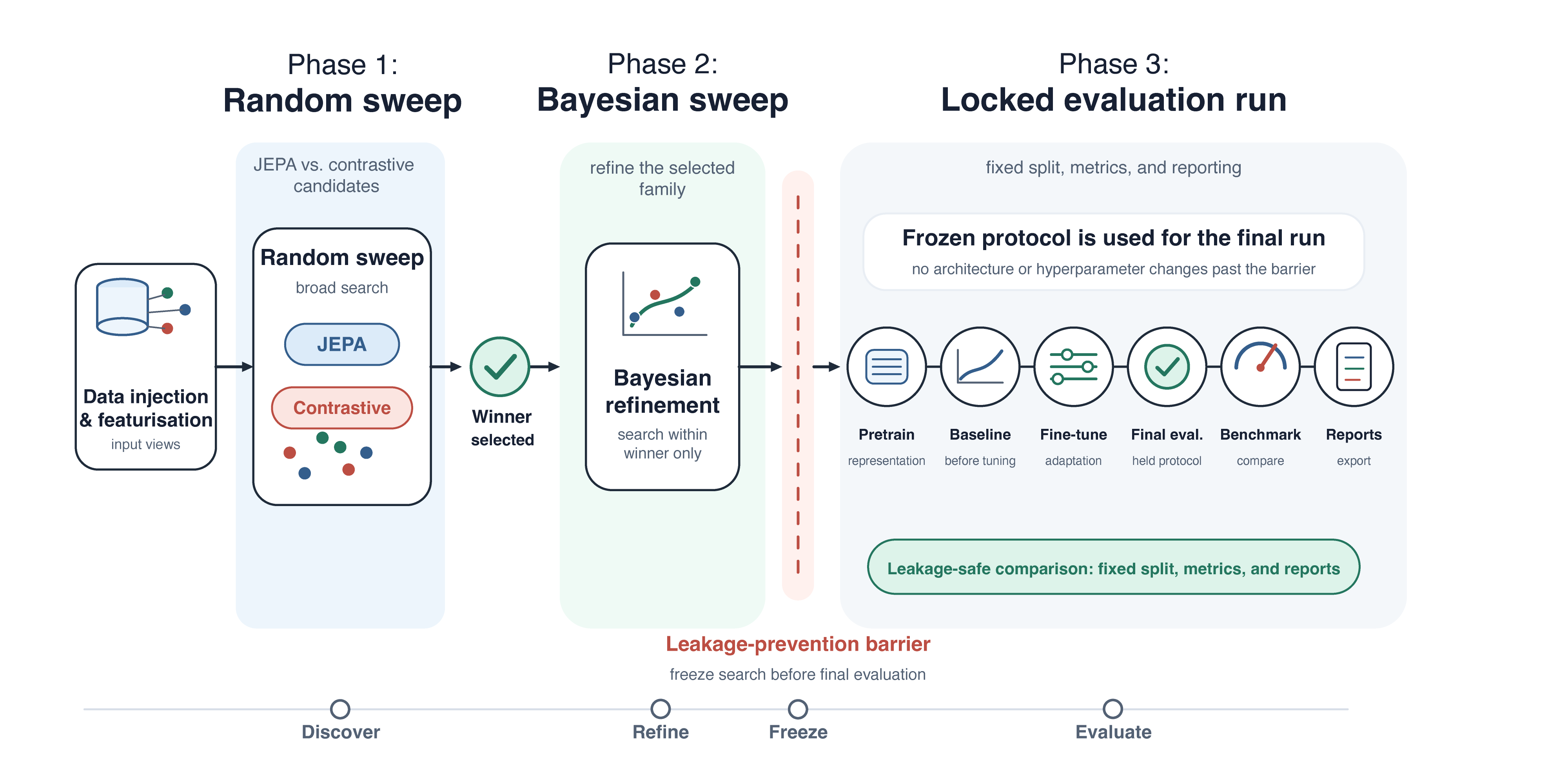}
\caption{Leakage-resistant, staged evaluation pipeline separating
objective-family choice (Phase 1), configuration selection (Phase 2),
and downstream scaffold-shift evaluation (Phase 3).}
\end{figure*}

\section{Results and Discussion}

We report results in three phases designed to reduce leakage and
selection bias. Phase 1 compares predictive self-supervision (JEPA)
against a compute-matched contrastive baseline across 80
objective-screening runs on ESOL, and quantifies the robustness of each
objective to masking and augmentation choices. Phase 2 conducts a
JEPA-only sweep to identify a stable `default' configuration for
downstream use. Phase 3 evaluates the selected configuration on Tox21
under Bemis--Murcko scaffold splits, with joint discrimination and
calibration analysis and motif-level attribution diagnostics.

Under identical hardware and training settings, JEPA runs were typically
shorter per run (≈15--25 min) than the contrastive baseline (≈35--45
min) (Supporting Information Table S2), consistent with the reduced
overhead from avoiding negative sampling and extensive augmentation
pipelines.

Phase 2 shows that augmentation choice is a dominant driver of
objective-screening performance: minimal augmentation (`none' or
`graph\_light') yields a clear lower-error band, whereas adding
geometric perturbations (`geom\_only' or `mixed\_light') substantially
degrades ESOL proxy RMSE. These trends motivate selecting a
light-augmentation, moderate-mask-ratio configuration for Phase 3.

On Tox21, improvements in ROC-AUC and PR-AUC are assay dependent (Table
4). Calibration metrics reveal that better ranking does not always imply
better probability estimates under scaffold shift; however, several
endpoints show simultaneous gains in discrimination and calibration, and
reliability diagrams highlight where hybrid fine-tuning improves (or
worsens) confidence alignment. Attribution maps provide additional
qualitative diagnostics by localizing model sensitivity to chemically
plausible substructures in selected examples.

\subsection{Phase 1 Objective
Screening}

Phase 1 screens the pretraining objective under strict compute caps and
without using downstream labels for selection. We sample 80
configurations spanning backbone, mask ratio, contiguity, 3D features,
and augmentation profile (Table 1), and compare JEPA to an InfoNCE
contrastive baseline using ESOL validation RMSE as an
objective-screening signal (Table 2).

JEPA yields lower RMSE across all matched backbone×seed pairs
(Supporting Information Table S3), with a mean paired ΔRMSE of 1.61 and
a 95\% bootstrap CI {[}1.52, 1.72{]}. A one-sided Wilcoxon signed-rank
test across pairs gives p = 2.4×10−4, supporting the qualitative
separation visible in the RMSE distributions (Figure 3). Importantly,
ESOL is used here as a proxy for objective selection---not as a claim
that solubility RMSE directly predicts toxicity classification---and
Phase 3 provides the downstream test under scaffold shift.

Under identical hardware and training settings, JEPA runs were typically
shorter per run (≈15--25 min) than the contrastive baseline (≈35--45
min) (Supporting Information Table S2), consistent with the reduced
overhead from avoiding negative sampling and extensive augmentation
pipelines.

\begin{table*}[tp]
\centering
\footnotesize
\renewcommand{\arraystretch}{1.35}\setlength{\tabcolsep}{3pt}
\caption{Phase-1 sweep configuration (paired, compute-capped ESOL proxy).}
\begin{tabular}{@{}>{\raggedright\arraybackslash}p{0.1500\textwidth}>{\raggedright\arraybackslash}p{0.2667\textwidth}>{\raggedright\arraybackslash}p{0.2667\textwidth}>{\raggedright\arraybackslash}p{0.2667\textwidth}@{}}
\toprule
\textbf{Group} & \textbf{Parameter} & \textbf{Value(s) in Phase-1} & \textbf{Notes} \\
\midrule\addlinespace[2pt]

Input \& data & Unlabelled corpus & ZINC (canonicalised) & Cached graph
dataset used for pretraining. \\
& Labelled proxy task & ESOL (regression) & Target: measured log
solubility (``measured log solubility in mols per litre''). \\
& Labelled sample size & 5,000 & Used only for proxy evaluation
stage. \\
& Unlabelled sampling cap & 100,000 sampled; 250,000 max graphs/run &
Constrains Phase-1 compute while keeping the objective comparison
meaningful. \\
Architecture & Backbone (gnn\_type) & \{GINE, D-MPNN, SchNet3D\} & Same
candidate sets both objectives. \\
& Depth / width & 3 layers; 256 hidden & Held constantly to isolate
objective effects. \\
Masking \&\\
view construction\strut & Contiguity & 1 (connected subgraph mask) & Implements
§4.1.2 ``chemically coherent fragment'' masking. \\
& Mask ratio ρ & JEPA: \{0.10, 0.15, 0.20, 0.25\}; Contrastive: fixed
0.15 & JEPA explicitly probes masking difficulty; contrastive sweep kept
ρ fixed. \\
& Augmentations & all enabled & rotate, bond-angle masking, dihedral
perturbation, bond deletion, atom masking, subgraph removal (all =
1). \\
Training dynamics & Pretraining & 5 epochs; batch size 256; LR 1e-4 &
Fixed across runs. \\
& Proxy stage & 10 epochs; batch size 256 & Fixed across runs. \\
& JEPA teacher EMA & \{0.95, 0.98, 0.99\} & Training-stability lever
specific to JEPA. \\
& Contrastive temperature & \{0.05, 0.10\} & Contrastive-specific
optimisation knob. \\
Reproducibility & Seeds & \{1,2,3,4,5\} & Seed used as a primary
matching key in paired effects. \\
\bottomrule
\end{tabular}
\end{table*}

\begin{table*}[tp]
\centering
\footnotesize
\renewcommand{\arraystretch}{1.35}\setlength{\tabcolsep}{3pt}
\caption{ESOL validation RMSE distributions (proxy), Phase-1 sweeps (n=80).}
\begin{tabular}{@{}>{\raggedright\arraybackslash}p{0.1500\textwidth}>{\raggedright\arraybackslash}p{0.1333\textwidth}>{\raggedright\arraybackslash}p{0.1333\textwidth}>{\raggedright\arraybackslash}p{0.1333\textwidth}>{\raggedright\arraybackslash}p{0.1333\textwidth}>{\raggedright\arraybackslash}p{0.1333\textwidth}>{\raggedright\arraybackslash}p{0.1333\textwidth}@{}}
\toprule
\textbf{Objective family} & \textbf{Runs} & \textbf{Mean RMSE} & \textbf{Std} & \textbf{Median} & \textbf{Best} & \textbf{Worst} \\
\midrule\addlinespace[2pt]

JEPA & 40 & 2.117 & 0.121 & 2.113 & 1.886 & 2.386 \\
Contrastive & 40 & 3.746 & 0.195 & 3.717 & 3.401 & 4.268 \\
\bottomrule
\end{tabular}
\end{table*}

\begin{figure*}[tp]
\centering
\includegraphics[width=0.92\textwidth]{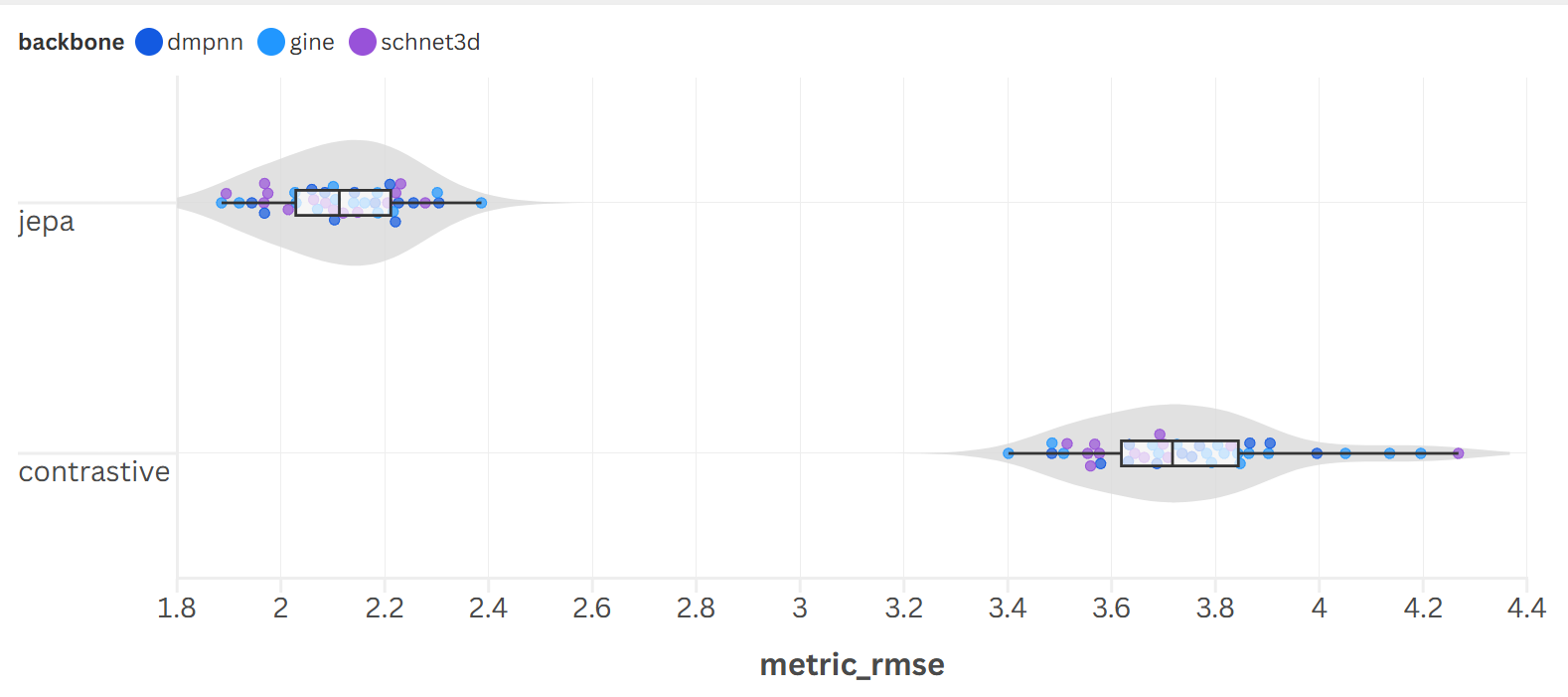}
\caption{Phase-1 objective comparison under matched compute (80
runs). M-JEPA achieves substantially lower ESOL proxy RMSE than the
compute-matched contrastive baseline.}
\end{figure*}

\subsection{Phase 2 Configuration
Selection}

Phase 2 performs a JEPA-only sweep to identify a stable default
configuration for downstream evaluation. We randomize across backbones,
mask ratios, contiguity, 3D features, learning rates, EMA decay, and
augmentation profiles, and rank configurations by ESOL validation RMSE
(Table 3).

Across the top-ranked configurations, performance concentrates in the
minimal-augmentation regime (none or graph\_light), with moderate mask
ratios (≈0.10--0.15) and EMA decay in the 0.95--0.98 range. Notably,
geometry-only or mixed geometric augmentations do not appear among the
best configurations, consistent with the Phase-2 RMSE bands shown in
Figure 4.

These findings suggest that, for molecular graphs, `heavier' view
corruption can remove chemically meaningful signal rather than
encouraging beneficial invariances. We therefore select the
best-performing light-augmentation configuration for Phase 3, and we
treat augmentation sensitivity as a key methodological consideration
when transferring SSL practices from vision to chemistry. (Wang et al.,
2021; Li et al., 2021; Wu et al., 2023).

This `augmentation paradox' likely reflects the fact that, unlike
images, small geometric perturbations in molecular graphs can distort
stereochemical and conformation-dependent cues---or even create
physically implausible geometries---thereby injecting `chemical noise'
that obscures the latent grammar of molecular structure. In contrast,
light graph-based augmentations may better preserve chemical validity
while still preventing trivial memorization. Practically, these results
suggest that molecular SSL may benefit more from conservative,
chemically informed discrete edits than from aggressive continuous
geometric jitter, especially when the goal is robust generalization
under scaffold shift.

For completeness, a representation-only transfer sanity check---freezing
the selected pretrained encoder and training a fresh regression head on
ESOL across multiple random seeds---is reported in the Supporting
Information (Section S6, Table S10) and is not used to support the main
claims.

\begin{figure*}[tp]
\centering
\includegraphics[width=0.92\textwidth]{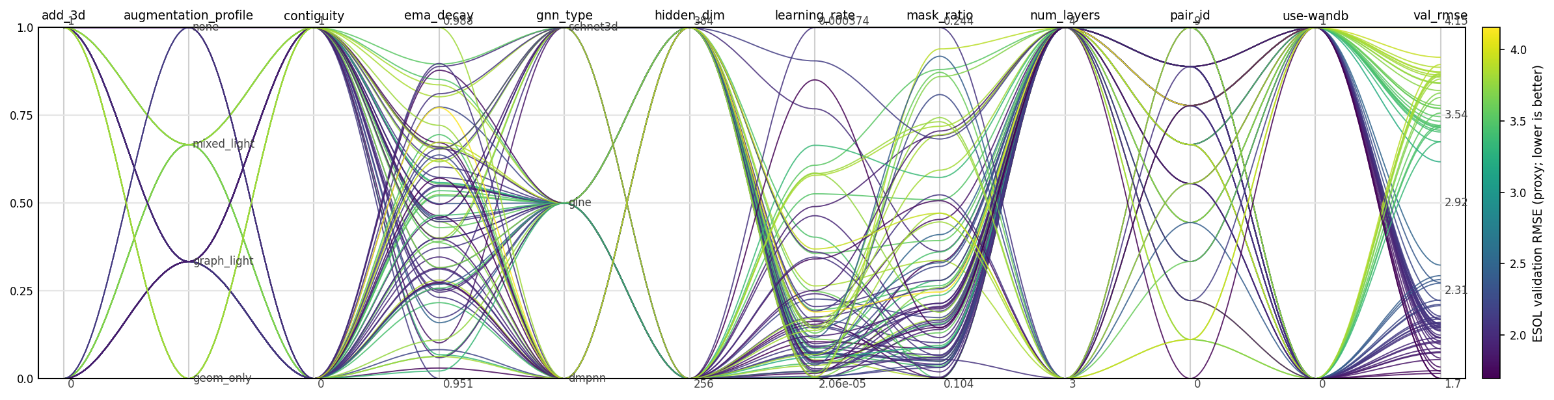}
\caption{Phase-2 sweep structure (parallel-coordinates view of
the explored hyperparameter space and outcomes).}
\end{figure*}

\begin{figure*}[tp]
\centering
\includegraphics[width=0.92\textwidth]{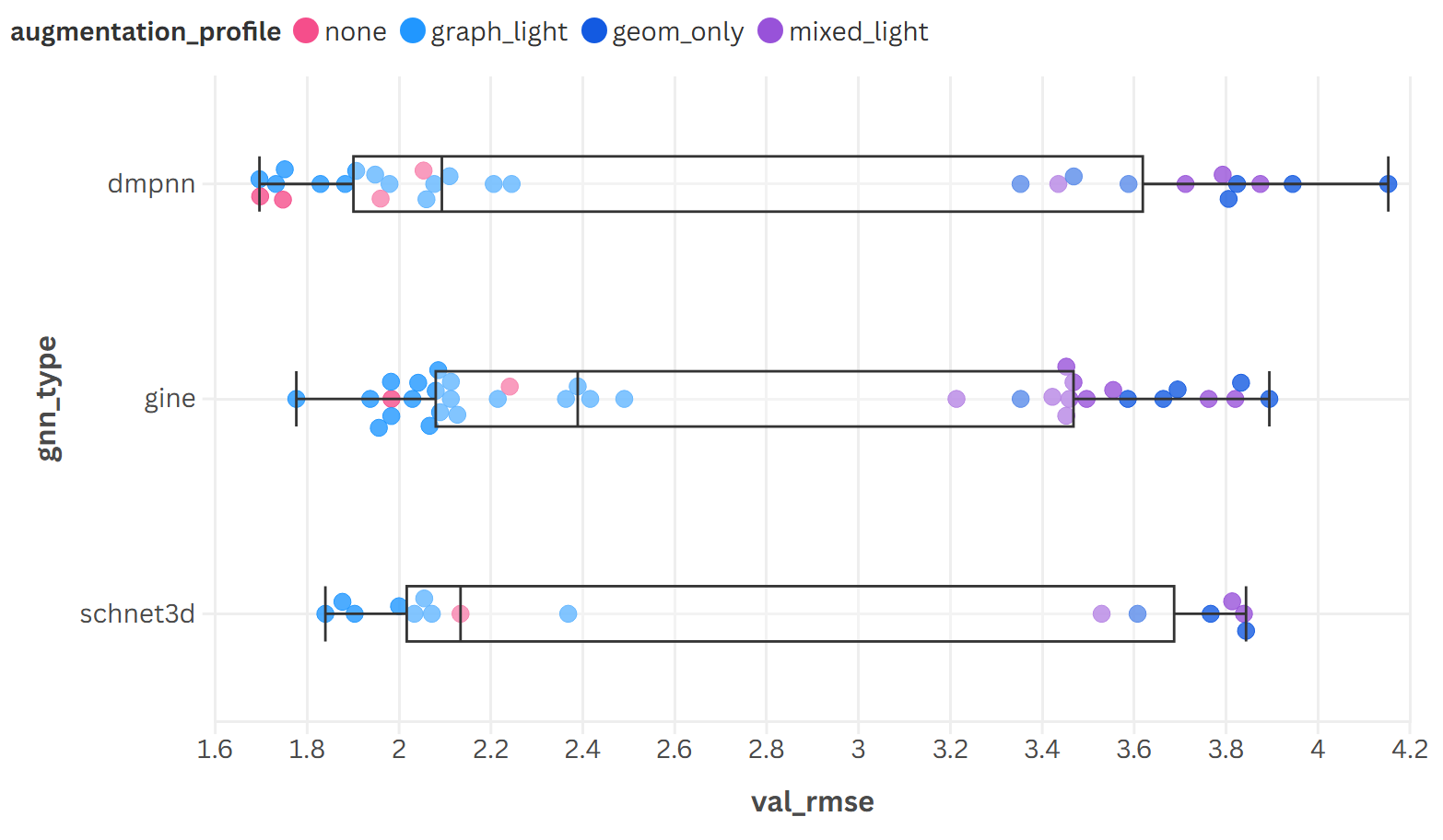}
\caption{ESOL proxy RMSE by augmentation profile and backbone
(Phase-2 sweep). Heavy geometric augmentation degrades performance
relative to minimal/no augmentation.}
\end{figure*}

\begin{figure*}[tp]
\centering
\includegraphics[width=0.92\textwidth]{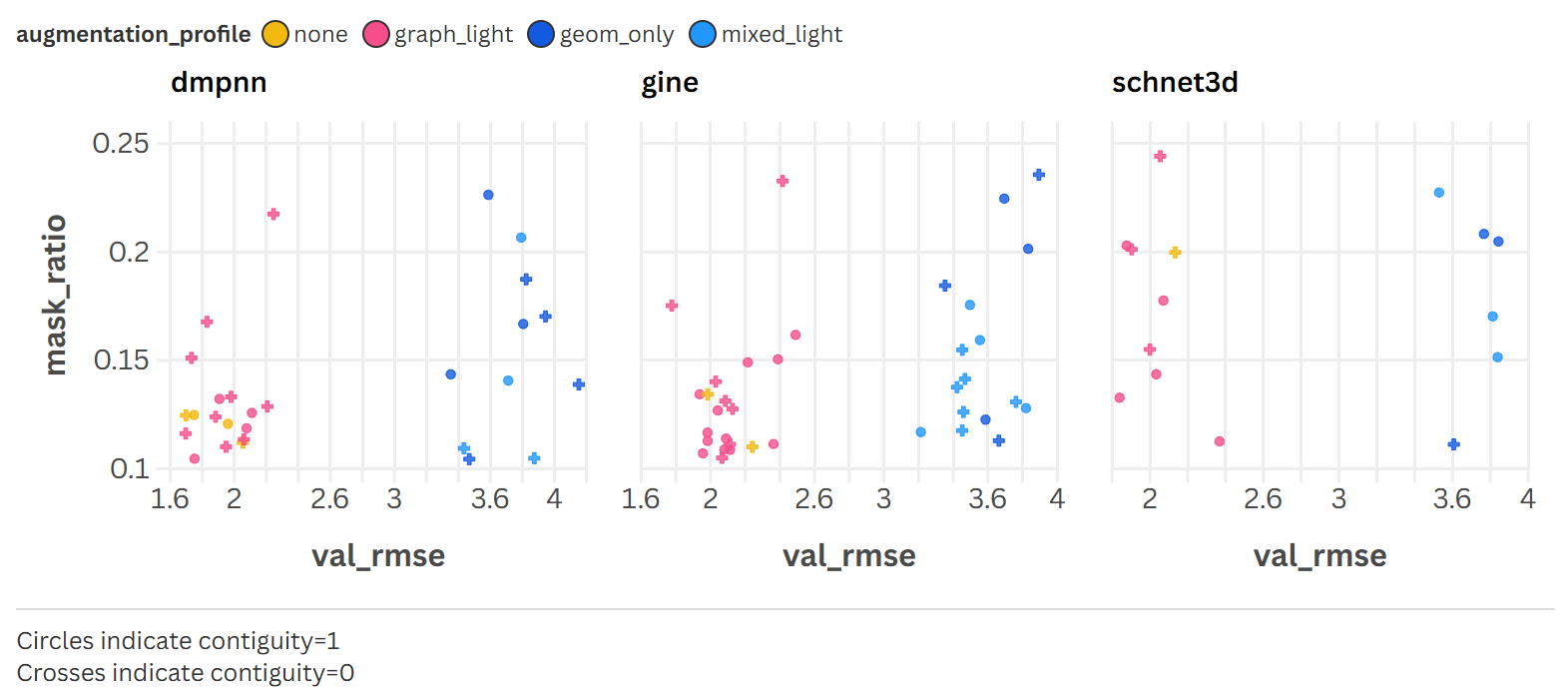}
\caption{Augmentation and masking effects across Phase-2 runs,
highlighting regime-level performance differences.}
\end{figure*}

\begin{sidewaystable*}[tp]
\centering
\scriptsize
\renewcommand{\arraystretch}{1.35}\setlength{\tabcolsep}{3pt}
\caption{Phase-2 seed recheck (top candidates rerun with extra seeds; n=3 per configuration).}
\resizebox{0.95\textheight}{!}{%
\begin{tabular}{@{}
  >{\arraybackslash}p{0.0737\textwidth}
  >{\arraybackslash}p{0.0890\textwidth}
  >{\arraybackslash}p{0.0742\textwidth}
  >{\arraybackslash}p{0.0344\textwidth}
  >{\arraybackslash}p{0.0640\textwidth}
  >{\arraybackslash}p{0.1152\textwidth}
  >{\arraybackslash}p{0.0639\textwidth}
  >{\arraybackslash}p{0.0897\textwidth}
  >{\arraybackslash}p{0.0512\textwidth}
  >{\arraybackslash}p{0.0768\textwidth}
  >{\arraybackslash}p{0.1181\textwidth}@{}}
\toprule
\textbf{Rank} & \textbf{Backbone} & \textbf{hid / layers} & \textbf{3d} & \textbf{contig} & \textbf{Aug} & \textbf{Mask ratio} & \textbf{lr} & \textbf{ema} & \textbf{Mean RMSE} & \textbf{95\% CI} \\
\midrule\addlinespace[2pt]

1 & D-MPNN & 256 / 4 & 0 & 0 & graph\_light & 0.151 & 2.57e-05 & 0.952 &
1.719 & {[}1.694, 1.744{]} \\
2 & D-MPNN & 384 / 4 & 1 & 1 & graph\_light & 0.105 & 7.82e-05 & 0.968 &
1.897 & {[}1.681, 2.012{]} \\
3 & D-MPNN & 256 / 4 & 1 & 0 & none & 0.125 & 3.21e-04 & 0.972 & 1.941 &
{[}1.802, 2.071{]} \\
4 & D-MPNN & 256 / 4 & 0 & 0 & graph\_light & 0.116 & 5.77e-05 & 0.961 &
1.942 & {[}1.797, 2.077{]} \\
5 & D-MPNN & 384 / 4 & 1 & 1 & none & 0.105 & 3.85e-05 & 0.976 & 2.015 &
{[}1.953, 2.070{]} \\
\bottomrule
\end{tabular}}
\end{sidewaystable*}

\subsection{Tox21 Scaffold-Split
Evaluation}

Assay-local discrimination and reliability metrics are reported in Table
4, and reliability behavior is visualized in Figures 7--9. Direct
comparison to published Tox21 results is complicated by differences in
scaffold partitioning, preprocessing, and tuning; for context,
Supporting Information Tables S6--S7 summarize representative
scaffold-split ROC-AUC ranges reported for common baselines. (Wu et al.,
2017; Mayr et al., 2016; Born et al., 2023).

Across assays, hybrid fine‑tuning improves average discrimination while
remaining assay‑dependent: mean ROC‑AUC increases from 0.546 (base) to
0.609 (hybrid) and mean PR‑AUC from 0.220 to 0.246. Gains are strongest
for several nuclear‑receptor endpoints (e.g., NR‑Aromatase,
NR‑ER/ER‑LBD) and stress‑response endpoints (e.g., SR‑ATAD5), while a
minority of assays show negligible or negative changes, emphasizing that
representation transfer does not affect all toxicological mechanisms
equally.

Calibration does not automatically track discrimination: on average the
hybrid model reduces the Brier score (0.088 → 0.076) and ECE (0.279 →
0.190), but several assays still exhibit miscalibration under scaffold
shift (Table 4; Figures 7--9). This divergence motivates reporting
discrimination and reliability jointly, and suggests that downstream
calibration strategies may be endpoint‑specific for highly imbalanced
assays These reliability metrics are reported as diagnostics of
probability quality under shift, not as proof of downstream decision
utility. (Brier, 1950; Guo et al., 2017).

Table 4. Assay-wise discrimination and calibration on Tox21 under
Bemis--Murcko scaffold splits. Base: supervised-from-scratch; Hybrid:
JEPA-initialized. Values are means over the evaluation ensemble; full
mean±SD and Δ metrics are provided in the Supporting Information.

\begin{sidewaystable*}[tp]
\centering
\scriptsize
\renewcommand{\arraystretch}{1.35}\setlength{\tabcolsep}{3pt}
\caption{}
\resizebox{0.95\textheight}{!}{%
\begin{tabular}{@{}
  >{\arraybackslash}p{0.0989\textwidth}
  >{\arraybackslash}p{0.0646\textwidth}
  >{\arraybackslash}p{0.0847\textwidth}
  >{\arraybackslash}p{0.0870\textwidth}
  >{\arraybackslash}p{0.0847\textwidth}
  >{\arraybackslash}p{0.0870\textwidth}
  >{\arraybackslash}p{0.0847\textwidth}
  >{\arraybackslash}p{0.0870\textwidth}
  >{\arraybackslash}p{0.0847\textwidth}
  >{\arraybackslash}p{0.0870\textwidth}@{}}
\toprule
\textbf{Assay} & \textbf{+ve Frac} & \textbf{ROC-AUC (Base)} & \textbf{ROC-AUC (Hybrid)} & \textbf{PR-AUC (Base)} & \textbf{PR-AUC (Hybrid)} & \textbf{Brier (Base)} & \textbf{Brier (Hybrid)} & \textbf{ECE (Base)} & \textbf{ECE (Hybrid)} \\
\midrule\addlinespace[2pt]

NR-AR & 0.0425 & 0.545 & 0.625 & 0.078 & 0.118 & 0.040 & 0.040 & 0.0149
& 0.0107 \\
NR-AR-LBD & 0.0351 & 0.678 & 0.696 & 0.081 & 0.098 & 0.031 & 0.028 &
0.0100 & 0.0276 \\
NR-AhR & 0.1173 & 0.626 & 0.739 & 0.212 & 0.312 & 0.258 & 0.395 & 0.363
& 0.153 \\
NR-Aromatase & 0.0515 & 0.477 & 0.561 & 0.103 & 0.143 & 0.253 & 0.089 &
0.406 & 0.025 \\
NR-ER & 0.1280 & 0.499 & 0.569 & 0.170 & 0.189 & 0.257 & 0.252 & 0.361 &
0.220 \\
NR-ER-LBD & 0.0503 & 0.534 & 0.545 & 0.069 & 0.061 & 0.180 & 0.041 &
0.372 & 0.001 \\
NR-PPAR-γ & 0.0288 & 0.427 & 0.635 & 0.053 & 0.085 & 0.057 & 0.057 &
0.022 & 0.022 \\
SR-ARE & 0.1615 & 0.526 & 0.532 & 0.240 & 0.244 & 0.254 & 0.213 & 0.275
& 0.185 \\
SR-ATAD5 & 0.0373 & 0.571 & 0.674 & 0.059 & 0.082 & 0.255 & 0.052 &
0.456 & 0.021 \\
SR-HSE & 0.0575 & 0.505 & 0.556 & 0.098 & 0.122 & 0.252 & 0.077 & 0.427
& 0.077 \\
SR-MMP & 0.1580 & 0.537 & 0.554 & 0.259 & 0.272 & 0.255 & 0.237 & 0.299
& 0.266 \\
SR-p53 & 0.0624 & 0.554 & 0.553 & 0.125 & 0.134 & 0.224 & 0.163 & 0.346
& 0.097 \\
\bottomrule
\end{tabular}}
\end{sidewaystable*}

\begin{figure}[tp]
\centering
\includegraphics[width=\columnwidth]{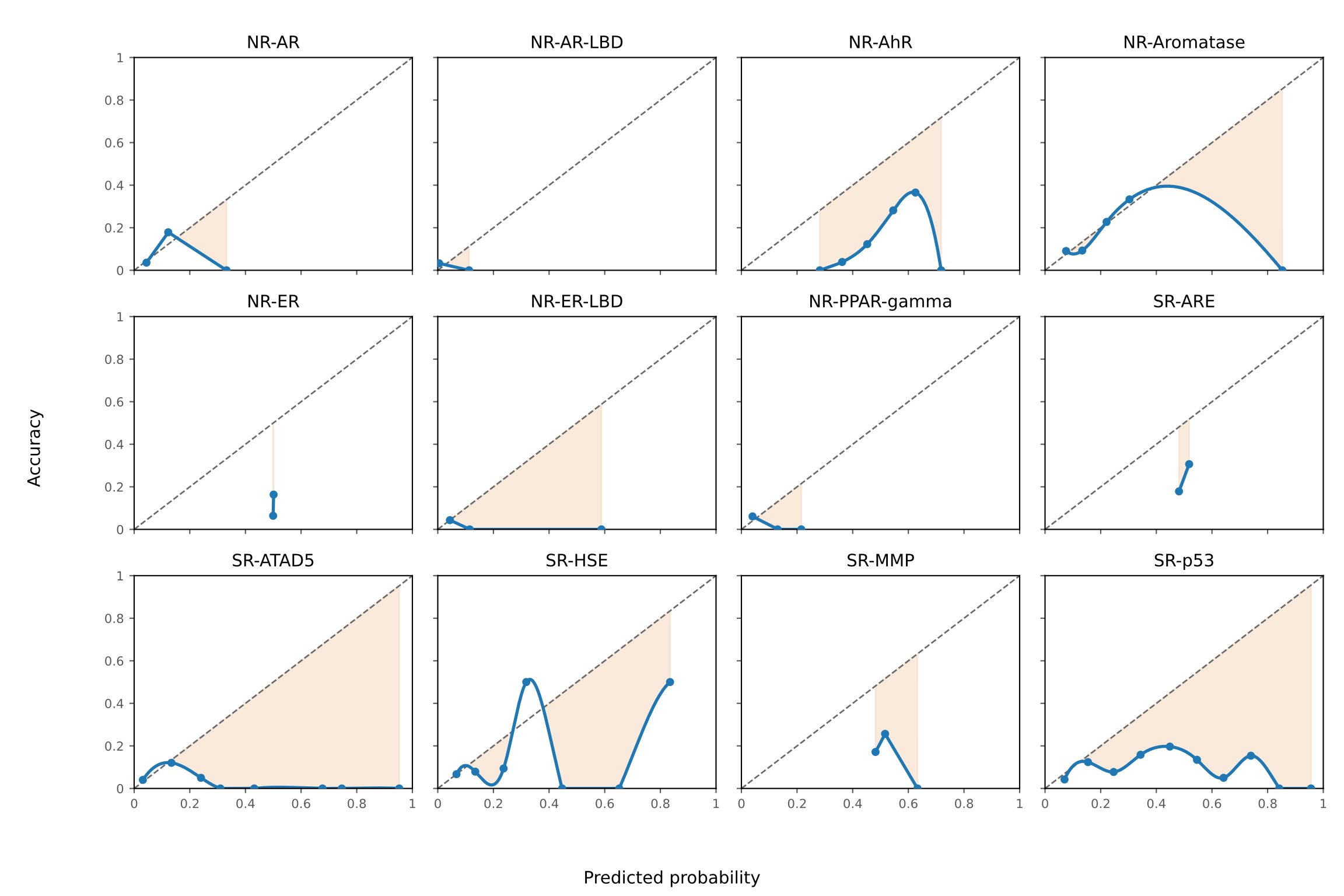}
\caption{Reliability diagrams for Tox21 assays under scaffold
split, comparing baseline (frozen) and hybrid fine-tuning settings.}
\end{figure}

\begin{figure}[tp]
\centering
\includegraphics[width=\columnwidth]{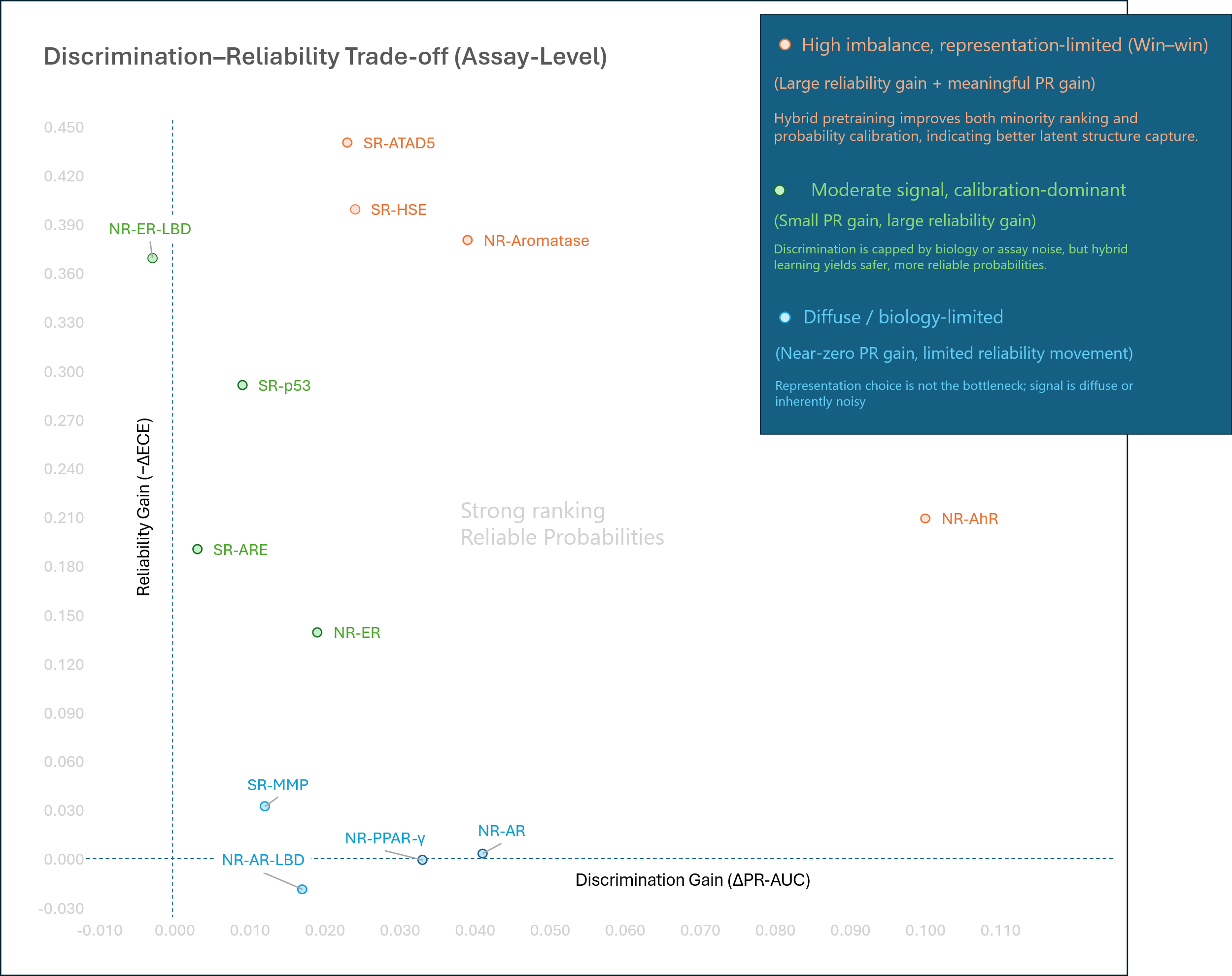}
\caption{Assay-level discrimination--reliability trade-off under
scaffold shift, plotted as ΔPR-AUC versus ΔECE from baseline to hybrid
fine-tuning.}
\end{figure}

\subsection{Attribution Diagnostics}

We complement discrimination and calibration metrics with motif‑level
attribution diagnostics to probe what chemical substructures the models
rely on under scaffold shift. We compute Integrated Gradients (IG) on
the assay‑specific output logit with respect to atom and bond features,
using a zero (masked) feature baseline, and aggregate per‑atom
attributions to chemically meaningful motifs (ring systems and
Bemis--Murcko scaffold components). Full attribution protocol and
example‑selection criteria are provided in the Supporting Information.
(Sundararajan et al., 2017; Sanchez-Lengeling et al., 2020; Ying et al.,
2019).

Across representative example molecules, motif-level attribution
rankings are often stable across fine-tuning checkpoints, but stability
is not guaranteed. We quantify this by correlating motif-rankings
between early and late checkpoints on three illustrative examples
(Spearman ρ = 0.95, 0.91, and 0.47; Supporting Information Table S12),
and we therefore treat attribution maps as diagnostic tools rather than
ground-truth explanations.

These analyses are intended as interpretability diagnostics rather than
validated toxicophore discovery; quantitative agreement with curated
alert libraries and broader toxicophore benchmarking are natural next
steps, but would require additional external resources beyond the scope
of the present study.

\begin{figure}[tp]
\centering
\includegraphics[width=\columnwidth]{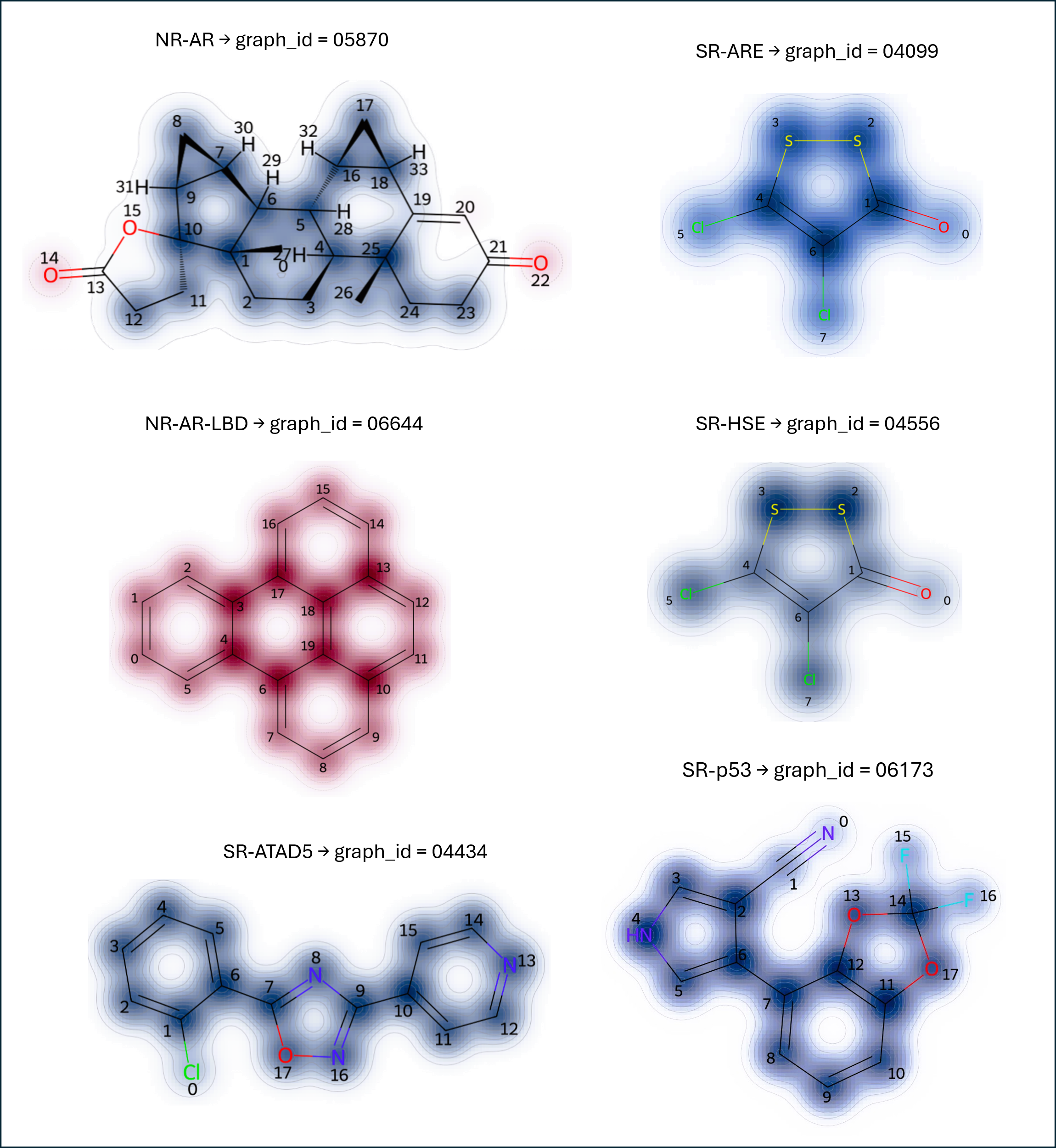}
\caption{Representative integrated-gradient attribution patterns
across Tox21 assays under scaffold shift.}
\end{figure}

\section{Conclusions}

We introduced M-JEPA, a predictive self-supervised objective for
molecular graphs based on connected-subgraph masking and an EMA teacher,
and evaluated it with a leakage-resistant, compute-matched pipeline that
separates objective screening from downstream testing. Across
compute-capped sweeps, JEPA is robust and consistently outperforms a
compute-matched contrastive baseline on ESOL proxy RMSE while also
reducing wall-clock runtime. On Tox21 scaffold splits, hybrid
fine-tuning yields assay-dependent gains and highlights that
discrimination improvements do not guarantee calibrated probabilities;
reliability metrics and diagrams should accompany AUC reporting for
safety-relevant use. A key limitation is baseline breadth: we focus on a
carefully compute-matched InfoNCE baseline rather than an exhaustive
comparison to all recent molecular SSL objectives, and additional
ablations (e.g., EMA removal, mask-ratio sensitivity) would further
isolate which components drive gains. Code and configuration files are
available at https://github.com/Kar488/M-JEPA/ (currently private;
available to reviewers upon request and intended for public release).

The Supporting Information provides detailed configurations, compute
environment, additional tables, and validation probes for graph
construction.

\section{Data and Code Availability}

All datasets used in this study are publicly available (ZINC, ESOL,
Tox21). The implementation, experiment configurations, and analysis
scripts are available at https://github.com/Kar488/M-JEPA/ (private at
the time of writing; access can be provided to reviewers upon request;
will be made public upon acceptance). For transparency, the full Phase‑1
objective screen sweep logs (80 runs; including run‑level configurations
and metric histories) are archived alongside the codebase.

\section{Author Contributions}

K. I. conceived the study, implemented the codebase, conducted all
experiments and analyses, and wrote the manuscript. N. S. provided
supervisory oversight and editorial feedback.

\section{Supporting Information}

Supporting Information is available and includes: compute environment
and package versions; additional sweep details; SOTA reference tables;
graph-construction validation probes; downstream Tox21 hybrid
configuration tables; and a representation-only transfer benchmark.

\section{References}

Assran, M., Duval, Q., Misra, I., Bojanowski, P., Vincent, P., Rabbat,
M., and Ballas, N. 2023. Self-Supervised Learning from Images with a
Joint Embedding Predictive Architecture IEEE/CVF Conference on Computer
Vision and Pattern Recognition (CVPR), Vancouver,BC,Canada.
https://ieeexplore.ieee.org/document/10205476

Bemis, G. W., and Murcko, M. A. 1996. The Properties of Known Drugs. 1.
Molecular Frameworks. Journal of Medicinal Chemistry, 39(15), 2887-2893.
https://doi.org/10.1021/jm9602928

Born, J., Markert, G., Janakarajan, N., Kimber, T. B., Volkamer, A.,
Martínez, M. R., and Manica, M. 2023. Chemical representation learning
for toxicity prediction doi: 10.1039/D2DD00099G. Digital Discovery,
2(3), 674--691. https://doi.org/10.1039/D2DD00099G

Brier, G. W. 1950. Verification of Forecasts Expressed in Terms of
Probability. Monthly Weather Review, 78(1), 1-3.
https://doi.org/10.1175/1520-0493(1950)078\textless0001:VOFEIT\textgreater2.0.CO;2

Chen, M., Radford, A., Child, R., Wu, J., Jun, H., Luan, D., and
Sutskever, I. 2020. Generative pretraining from pixels Proceedings of
the 37th International Conference on Machine Learning,

Delaney, J. S. 2004. ESOL: Estimating Aqueous Solubility Directly from
Molecular Structure. Journal of Chemical Information and Computer
Sciences, 44(3), 1000-1005. https://doi.org/10.1021/ci034243x

Deng, J., Yang, Z., Wang, H., Ojima, I., Samaras, D., and Wang, F. 2023.
A systematic study of key elements underlying molecular property
prediction. Nature Communications, 14(1), 6395.
https://doi.org/10.1038/s41467-023-41948-6

Ebner, A., Bartmann, C., Luukkonen, S., Schimunek, J., and Klambauer, G.
2025. Measuring AI Progress in Drug Discovery: A Reproducible
Leaderboard for the Tox21 Challenge. https://arxiv.org/abs/2511.14744

Ellington, C., Li, D., Zou, S., Cole, E., Sun, N., Addagudi, S., and
Xing, E. 2025. Rapid and Reproducible Multimodal Biological Foundation
Model Development with AIDO.ModelGenerator. Obesity, Fitness and
Wellness Week, 95. https://doi.org/10.1101/2025.06.30.662437

Gilmer, J., Schoenholz, S. S., Riley, P. F., Vinyals, O., and Dahl, G.
E. 2017. Neural Message Passing for Quantum Chemistry. In Proceedings of
the 34th International Conference on Machine Learning (ICML), PMLR 70,
1263--1272. https://arxiv.org/abs/1704.01212

Guo, C., Pleiss, G., Sun, Y., and Weinberger, K. Q. 2017. On Calibration
of Modern Neural Networks. Proceedings of the 34th International
Conference on Machine Learning (ICML). https://arxiv.org/abs/1706.04599

Hirohara, M., Saito, Y., Koda, Y., Sato, K., and Sakakibara, Y. 2018.
Convolutional neural network based on SMILES representation of compounds
for detecting chemical motif. BMC Bioinformatics, 19(19), 526.
https://doi.org/10.1186/s12859-018-2523-5

Hu, W., Fey, M., Zitnik, M., Dong, Y., Ren, H., Liu, B., Catasta, M.,
and Leskovec, J. 2020. Open graph benchmark: datasets for machine
learning on graphs Proceedings of the 34th International Conference on
Neural Information Processing Systems, Vancouver, BC, Canada.

Hu, W., Liu, B., Gomes, J., Zitnik, M., Liang, P., Pande, V. S., and
Leskovec, J. 2020. Strategies for Pre-training Graph Neural Networks.
https://openreview.net/forum?id=HJlWWJSFDH

Huang, K., Chandak, P., Wang, Q., Havaldar, S., Vaid, A., Leskovec, J.,
and Zitnik, M. 2024. A foundation model for clinician-centred drug
repurposing (TxGNN). Nature Medicine, 30, 3601--3613.
https://doi.org/10.1038/s41591-024-03233-x

Idakwo, G., Thangapandian, S., Luttrell, J., Li, Y., Wang, N., Zhou, Z.,
Hong, H., Yang, B., Zhang, C., and Gong, P. 2020. Structure--activity
relationship-based chemical classification of highly imbalanced Tox21
datasets. Journal of Cheminformatics, 12(1), 66.
https://doi.org/10.1186/s13321-020-00468-x

Jeon, W., and Kim, D. 2019. FP2VEC: a new molecular featurizer for
learning molecular properties. Bioinformatics, 35(23), 4979--4985.
https://doi.org/10.1093/bioinformatics/btz307

Kang, L., Zhou, S., Fang, S., and Liu, S. 2024. Adapting differential
molecular representation with hierarchical prompts for multi-label
property prediction. Briefings in Bioinformatics, 25(5).
https://doi.org/10.1093/bib/bbae438

Kearnes, S., McCloskey, K., Berndl, M., Pande, V., and Riley, P. 2016.
Molecular graph convolutions: moving beyond fingerprints. Journal of
Computer-Aided Molecular Design, 30(8), 595--608.
https://doi.org/10.1007/s10822-016-9938-8

Landrum, G. 2016. RDKit: Open-Source Cheminformatics Software.
http://www.rdkit.org (accessed 2026-01-19).

Li, J., and Jiang, X. 2021. Mol-BERT: An Effective Molecular
Representation with BERT for Molecular Property Prediction. Wireless
Communications and Mobile Computing, 2021(1), 7181815.
https://doi.org/https://doi.org/10.1155/2021/7181815

Li, S., Zhou, J., Xu, T., Dou, D., and Xiong, H. 2021. GeomGCL:
Geometric Graph Contrastive Learning for Molecular Property Prediction
Thirty-Sixth AAAI Conference on Artificial Intelligence, Virtual.
https://ojs.aaai.org/index.php/AAAI/article/view/20377

Lim, S., and Lee, Y. O. (2021, 10--15 Jan. 2021). Predicting Chemical
Properties using Self-Attention Multi-task Learning based on SMILES
Representation. 2020 25th International Conference on Pattern
Recognition (ICPR),

Liyaqat, T., Ahmad, T., and Saxena, C. 2025. Advancements in Molecular
Property Prediction: A Survey of Single and Multimodal Approaches. Arch
Computat Methods Eng. https://doi.org/10.1007/s11831-025-10317-5

Mayr, A., Klambauer, G., Unterthiner, T., and Hochreiter, S. 2016.
DeepTox: Toxicity Prediction using Deep Learning {[}Original
Research{]}. Frontiers in Environmental Science, Volume 3 - 2015.
https://doi.org/10.3389/fenvs.2015.00080

Piccoli, F., Vogel, G., and Weber, J. 2025. Joint Embedding Predictive
Architecture for self-supervised pretraining on polymer molecular
graphs. https://arxiv.org/abs/2506.18194

Rong, Y., Bian, Y., Xu, T., Xie, W., Wei, Y., Huang, W., and Huang, J.
2020. Self-supervised graph transformer on large-scale molecular data
Proceedings of the 34th International Conference on Neural Information
Processing Systems, Vancouver, BC, Canada.

Sanchez-Lengeling, B., Wei, J., Lee, B., Reif, E., Wang, P. Y., Qian, W.
W., McCloskey, K., Colwell, L., and Wiltschko, A. 2020. Evaluating
attribution for graph neural networks Proceedings of the 34th
International Conference on Neural Information Processing Systems,
Vancouver, BC, Canada.

Schütt, K., Kindermans, P. J., Sauceda, H., Chmiela, S., Tkatchenko, A.,
and Müller, K. R. 2017. SchNet: A Continuous-Filter Convolutional Neural
Network for Modeling Quantum Interactions. Proceedings of the 31st
International Conference on Neural Information Processing Systems, Long
Beach, California, USA.

Segler, M. H., Kogej, T., Tyrchan, C., and Waller, M. P. 2018.
Generating Focused Molecule Libraries for Drug Discovery with Recurrent
Neural Networks. ACS Cent. Sci., 4(1), 120--131.
https://doi.org/10.1021/acscentsci.7b00512

Shen Wan, X., Zeng, X., Zhu, F., Wang Ya, l., Chu, Q., Tan, Y., Jiang
Yu, Y., and Chen Yu, Z. 2021. Out-of-the-box deep learning prediction of
pharmaceutical properties by broadly learned knowledge-based molecular
representations. Nature Machine Intelligence, 3(4), 334--343.
https://doi.org/10.1038/s42256-021-00301-6

Simpson, L., Millar, K., Cheng, A., Lim, C.-C., and Chew, H. G. 2025.
Graph-Based Integrated Gradients for\&nbsp;Explaining Graph Neural
Networks AI 2025: Advances in Artificial Intelligence: 38th Australasian
Joint Conference on Artificial Intelligence, AI 2025, Canberra, ACT,
Australia, December 1--5, 2025, Proceedings, Part I, Canberra, ACT,
Australia. https://doi.org/10.1007/978-981-95-4969-6\_12

Skenderi, G., Li, H., Tang, J., and Cristani, M. 2024. Graph-level
Representation Learning with Joint-Embedding Predictive Architectures.
https://openreview.net/forum?id=eddd0YTCiq

Soares, E., Shirasuna, V., Brazil, E., Cerqueira, R., Zubarev, D., and
Schmidt, K. 2024. A Large Encoder-Decoder Family of Foundation Models
for Chemical Language. https://arxiv.org/abs/2407.20267

Soleimany, A. P., Amini, A., Goldman, S., Rus, D., Bhatia, S. N., and
Coley, C. W. 2021. Evidential Deep Learning for Guided Molecular
Property Prediction and Discovery. ACS Central Science, 7(8),
1356--1367. https://doi.org/10.1021/acscentsci.1c00546

Sterling, T., and Irwin, J. J. 2015. ZINC 15 - Ligand Discovery for
Everyone. Journal of Chemical Information and Modeling, 55(11),
2324-2337. https://doi.org/10.1021/acs.jcim.5b00559

Stokes, J., Yang, K., Swanson, K., Wengong, J., Cubillos-Ruiz, A., and
Donghia, N. M. 2020. A Deep Learning Approach to Antibiotic Discovery.
Cell, 180(4), 688--702. https://doi.org/10.1016/j.cell.2020.01.021

Sundararajan, M., Taly, A., and Yan, Q. 2017. Axiomatic attribution for
deep networks Proceedings of the 34th International Conference on
Machine Learning - Volume 70, Sydney, NSW, Australia.

Wang, Y., Li, Z., and Barati Farimani, A. 2023. Graph Neural Networks
for Molecules. In C. Qu and H. Liu (Eds.), Machine Learning in Molecular
Sciences (pp. 21--66). Springer International Publishing.
https://doi.org/10.1007/978-3-031-37196-7\_2

Wang, Y., Yuan, Y., Li, Z., Zhang, Z., Ma, J., and Tang, J. 2021.
MolCLR: Molecular Contrastive Learning of Representations via Graph
Neural Networks. Nature Machine Intelligence, 4, 279--282.
https://doi.org/10.1038/s42256-022-00447-x

Withnall, M., Lindelöf, E., Engkvist, O., and Chen, H. 2020. Building
attention and edge message passing neural networks for bioactivity and
physical--chemical property prediction. Journal of Cheminformatics,
12(1), 1. https://doi.org/10.1186/s13321-019-0407-y

Wu, L., Lin, H., Tan, C., Gao, Z., and Li, S. Z. 2023. Self-Supervised
Learning on Graphs: Contrastive, Generative, or Predictive. IEEE
Transactions on Knowledge and Data Engineering, 35(4), 4216--4235.
https://doi.org/10.1109/TKDE.2021.3131584

Wu, Z., Jiang, D., Hsieh, C.-Y., Chen, G., Liao, B., Cao, D., and Hou,
T. 2021. Hyperbolic relational graph convolution networks plus: a simple
but highly efficient QSAR-modeling method. Briefings in Bioinformatics,
22(5). https://doi.org/10.1093/bib/bbab112

Wu, Z., Ramsundar, B., Feinberg, E. N., Gomes, J., Geniesse, C., Pappu,
A. S., and Pande, V. 2017. MoleculeNet: a benchmark for molecular
machine learning. Chemical Science. https://doi.org/10.1039/C7SC02664A

Xiong, Z., Wang, D., Liu, X., Zhong, F., Wan, X., Li, X., Li, Z., Luo,
X., Chen, K., Jiang, H., and Zheng, M. 2020. Pushing the Boundaries of
Molecular Representation for Drug Discovery with the Graph Attention
Mechanism. Journal of Medicinal Chemistry, 63(16), 8749--8760.
https://doi.org/10.1021/acs.jmedchem.9b00959

Xu, L., Xia, L., Pan, S., and Li, Z. 2024. Triple Generative
Self-Supervised Learning Method for Molecular Property Prediction.
International Journal of Molecular Sciences, 25, 3794.
https://doi.org/10.3390/ijms25073794

Yang, K., Swanson, K., Jin, W., Coley, C., Eiden, P., Gao, H.,
Guzman-Perez, A., Hopper, T., Kelley, B., Mathea, M., Palmer, A.,
Settels, V., Jaakkola, T., Jensen, K., and Barzilay, R. 2019. Analyzing
Learned Molecular Representations for Property Prediction. Journal of
Chemical Information and Modeling, 59(8), 3370--3388.
https://doi.org/10.1021/acs.jcim.9b00237

Ying, R., Bourgeois, D., You, J., Zitnik, M., and Leskovec, J. 2019.
GNNExplainer: generating explanations for graph neural networks. In
Proceedings of the 33rd International Conference on Neural Information
Processing Systems (pp. Article 829). Curran Associates Inc.

Yuan, Q., Szczypiński, F. T., and Jelfs, K. E. 2022. Explainable graph
neural networks for organic cages doi: 10.1039/D1DD00039J. Digital
Discovery, 1(2), 127--138. https://doi.org/10.1039/D1DD00039J

Zang, X., Zhao, X., and Tang, B. 2023. Hierarchical Molecular Graph
Self-Supervised Learning for property prediction. Communications
Chemistry, 6(1), 34. https://doi.org/10.1038/s42004-023-00825-5

Zeng, X., Xiang, H., Yu, L., Wang, J., Li, K., Nussinov, R., and Cheng,
F. 2022. Accurate prediction of molecular properties and drug targets
using a self-supervised image representation learning framework. Nature
Machine Intelligence, 4(11), 1004--1016.
https://doi.org/10.1038/s42256-022-00557-6

%% file: si.tex
\clearpage
\onecolumn
\begin{center}{\Large\bfseries\color{accent} Supporting Information}\\[4pt]{\normalsize M-JEPA: Predictive Self-Supervised Learning for Molecular Graphs}\end{center}\vspace{10pt}

\textbf{Supporting Information for: M-JEPA: Leakage-Resistant Evaluation
of Predictive Self-Supervised Learning for Small-Molecule Property
Prediction}

\textbf{Karthik Iyer}

\section{S1. Computation Environment and Reproducibility}

All experiments were executed in a container-isolated micromamba
environment with pinned Python and CUDA versions. GPU availability and
CUDA compatibility were validated at runtime (e.g., via nvidia-smi and
torch.cuda.is\_available()). Torch-Geometric extensions were installed
via pre-compiled CUDA wheels where possible, with source builds used
only as a controlled fallback.

Distributed Data Parallel (DDP) was intentionally disabled for
cache-warming and the Phase-1/Phase-2 sweeps. These stages consist of
many short, independent jobs with substantial disk I/O and frequent
experiment logging, where DDP introduces launch and synchronization
overhead without materially improving time-to-result.

All experiments use a single compute environment (Vast.ai instance);
reproducibility across different hardware is not tested.

\subsection{S1.1 Hardware Configuration}

{\centering
\small
\renewcommand{\arraystretch}{1.3}\setlength{\tabcolsep}{3pt}
\begin{longtable}{@{}>{\raggedright\arraybackslash}p{0.4500\linewidth}>{\raggedright\arraybackslash}p{0.4500\linewidth}@{}}
\caption*{\textbf{Table S1.} Hardware configuration for experiments.}\\
\toprule
\begin{minipage}[b]{\linewidth}\raggedright
\textbf{Component}
\end{minipage} & \begin{minipage}[b]{\linewidth}\raggedright
\textbf{Specification}
\end{minipage} \\
\midrule
\endfirsthead
\toprule
\begin{minipage}[b]{\linewidth}\raggedright
\textbf{Component}
\end{minipage} & \begin{minipage}[b]{\linewidth}\raggedright
\textbf{Specification}
\end{minipage} \\
\midrule
\endhead
Vast Compute Instance ID & 29371882 \\
Vast Host ID/Machine ID & 79986 / 14726 \\
GPU Architecture & NVIDIA GPU (CUDA-capable) \\
CUDA Capability & CUDA 12.6 (runtime), CUDA Toolkit 12.9 \\
GPU Compute Performance & 325.6 TFLOPS (FP16/TF32 aggregate,
provider-reported) \\
DLPerf Score & 169.4 DLPerf \\
GPU Memory (VRAM) & 96.0 GB total (1.8 GB used at capture time) \\
GPU Memory Bandwidth & 894.8 GB/s \\
CPU & AMD EPYC 7702, 64 cores \\
CPU Threads Available & 128 logical / 256 total \\
System Memory (RAM) & 1026 GB total (167 GB used at capture time) \\
Primary Storage & WD\_BLACK SN850X NVMe SSD \\
Disk Throughput & 10,444.8 MB/s \\
Disk Capacity & 1,045.5 GB total (91.0 GB used at capture time) \\
Motherboard & ROME2D32GM-2T \\
PCIe Interface & PCIe 4.0 ×16 \\
PCIe Bandwidth & 22.4 GB/s \\
Network Bandwidth & 5971.2 Mbps upload / 6601.7 Mbps download \\
Ports Available & 400 \\
\bottomrule
\end{longtable}\par}

\subsection{S1.2 Software Environment}

{\centering
\small
\renewcommand{\arraystretch}{1.3}\setlength{\tabcolsep}{3pt}
\begin{longtable}{@{}>{\raggedright\arraybackslash}p{0.1600\linewidth}>{\raggedright\arraybackslash}p{0.4100\linewidth}>{\raggedright\arraybackslash}p{0.4100\linewidth}@{}}
\caption*{\textbf{Table S2.} Key software packages and versions.}\\
\toprule
\begin{minipage}[b]{\linewidth}\raggedright
\textbf{Category}
\end{minipage} & \begin{minipage}[b]{\linewidth}\raggedright
\textbf{Package}
\end{minipage} & \begin{minipage}[b]{\linewidth}\raggedright
\textbf{Version / Notes}
\end{minipage} \\
\midrule
\endfirsthead
\toprule
\begin{minipage}[b]{\linewidth}\raggedright
\textbf{Category}
\end{minipage} & \begin{minipage}[b]{\linewidth}\raggedright
\textbf{Package}
\end{minipage} & \begin{minipage}[b]{\linewidth}\raggedright
\textbf{Version / Notes}
\end{minipage} \\
\midrule
\endhead
Deep Learning & PyTorch & 2.8.* (CUDA build, cu12x) \\
& TensorFlow & Installed (version per pip resolution) \\
Graph Learning & torch-geometric & 2.5.3 \\
& torch-scatter & 2.1.2 (wheel or source-built depending on CUDA
compatibility) \\
Scientific Computing & NumPy & \textless{} 2.0 \\
& SciPy & Installed via conda \\
& scikit-learn & Latest compatible \\
Chemoinformatics & RDKit & 2023.09.5 (conda) \\
& rdkit-pypi & 2022.9.5 (fallback for PyPI workflows) \\
Molecular Utilities & DeepChem & 2.8.0 \\
Reinforcement / Simulation & gymnasium & Installed \\
3-D Molecular Visuals & buildamol & Optional, GPU-accelerated \\
\bottomrule
\end{longtable}\par}

\section{S2. Additional Sweep Details}

This section provides additional tabular detail for objective screening
and Phase-2 configuration selection. These tables support
reproducibility and document the control knobs varied in the sweeps.

\subsection{S2.1 Phase-1 Paired Differences}

{\centering
\footnotesize
\renewcommand{\arraystretch}{1.3}\setlength{\tabcolsep}{3pt}
\begin{longtable}{@{}>{\raggedright\arraybackslash}p{0.1600\linewidth}>{\raggedright\arraybackslash}p{0.1367\linewidth}>{\raggedright\arraybackslash}p{0.1367\linewidth}>{\raggedright\arraybackslash}p{0.1367\linewidth}>{\raggedright\arraybackslash}p{0.1367\linewidth}>{\raggedright\arraybackslash}p{0.1367\linewidth}>{\raggedright\arraybackslash}p{0.1367\linewidth}@{}}
\caption*{\textbf{Table S3.} Paired objective comparison: ΔRMSE values (contrastive − M‑JEPA) across backbone/seed pairs. All Phase‑1 objective screening runs use the same fixed augmentation regime (all operators enabled) for both objectives.}\\
\toprule
\begin{minipage}[b]{\linewidth}\raggedright
\textbf{Backbone}
\end{minipage} & \begin{minipage}[b]{\linewidth}\raggedright
\textbf{Seed}
\end{minipage} & \begin{minipage}[b]{\linewidth}\raggedright
\textbf{n\\
(JEPA)}\strut
\end{minipage} & \begin{minipage}[b]{\linewidth}\raggedright
\textbf{mean RMSE\\
(JEPA)}\strut
\end{minipage} & \begin{minipage}[b]{\linewidth}\raggedright
\textbf{N\\
(Contrastive)}\strut
\end{minipage} & \begin{minipage}[b]{\linewidth}\raggedright
\textbf{mean RMSE\\
(Contrastive)}\strut
\end{minipage} & \begin{minipage}[b]{\linewidth}\raggedright
\textbf{Δ RMSE\\
(C − J)}\strut
\end{minipage} \\
\midrule
\endfirsthead
\toprule
\begin{minipage}[b]{\linewidth}\raggedright
\textbf{Backbone}
\end{minipage} & \begin{minipage}[b]{\linewidth}\raggedright
\textbf{Seed}
\end{minipage} & \begin{minipage}[b]{\linewidth}\raggedright
\textbf{n\\
(JEPA)}\strut
\end{minipage} & \begin{minipage}[b]{\linewidth}\raggedright
\textbf{mean RMSE\\
(JEPA)}\strut
\end{minipage} & \begin{minipage}[b]{\linewidth}\raggedright
\textbf{N\\
(Contrastive)}\strut
\end{minipage} & \begin{minipage}[b]{\linewidth}\raggedright
\textbf{mean RMSE\\
(Contrastive)}\strut
\end{minipage} & \begin{minipage}[b]{\linewidth}\raggedright
\textbf{Δ RMSE\\
(C − J)}\strut
\end{minipage} \\
\midrule
\endhead
dmpnn & 1 & 4 & 2.0987 & 2 & 3.7115 & 1.6128 \\
dmpnn & 2 & 3 & 2.2148 & 2 & 3.6197 & 1.4049 \\
dmpnn & 4 & 4 & 2.1084 & 3 & 3.8470 & 1.7386 \\
dmpnn & 5 & 1 & 2.2266 & 2 & 3.6330 & 1.4065 \\
gine & 2 & 1 & 1.8857 & 4 & 3.9393 & 2.0536 \\
gine & 3 & 4 & 2.2435 & 2 & 3.6389 & 1.3954 \\
gine & 4 & 7 & 2.0854 & 5 & 3.7801 & 1.6947 \\
gine & 5 & 2 & 2.1285 & 6 & 3.7648 & 1.6363 \\
schnet3d & 1 & 4 & 2.1084 & 1 & 3.6928 & 1.5844 \\
schnet3d & 3 & 5 & 2.1321 & 5 & 3.7209 & 1.5888 \\
schnet3d & 4 & 2 & 2.1264 & 2 & 3.6379 & 1.5114 \\
schnet3d & 5 & 3 & 1.9754 & 4 & 3.6761 & 1.7008 \\
\bottomrule
\end{longtable}\par}

\subsection{S2.2 Phase-2 Search Space}

Note: Augmentation profiles are referenced by short labels (none,
graph\_light, geom\_only, mixed\_light) to avoid truncation in tables;
Table S4 lists the corresponding binary toggles for each augmentation.

{\centering
\small
\renewcommand{\arraystretch}{1.3}\setlength{\tabcolsep}{3pt}
\begin{longtable}{@{}>{\raggedright\arraybackslash}p{0.1600\linewidth}>{\raggedright\arraybackslash}p{0.2733\linewidth}>{\raggedright\arraybackslash}p{0.2733\linewidth}>{\raggedright\arraybackslash}p{0.2733\linewidth}@{}}
\caption*{\textbf{Table S4.} Phase-2 configuration search space (M-JEPA).}\\
\toprule
\begin{minipage}[b]{\linewidth}\raggedright
\textbf{Group}
\end{minipage} & \begin{minipage}[b]{\linewidth}\raggedright
\textbf{Parameter}
\end{minipage} & \begin{minipage}[b]{\linewidth}\raggedright
\textbf{Value(s) explored in Phase 2}
\end{minipage} & \begin{minipage}[b]{\linewidth}\raggedright
\textbf{Notes}
\end{minipage} \\
\midrule
\endfirsthead
\toprule
\begin{minipage}[b]{\linewidth}\raggedright
\textbf{Group}
\end{minipage} & \begin{minipage}[b]{\linewidth}\raggedright
\textbf{Parameter}
\end{minipage} & \begin{minipage}[b]{\linewidth}\raggedright
\textbf{Value(s) explored in Phase 2}
\end{minipage} & \begin{minipage}[b]{\linewidth}\raggedright
\textbf{Notes}
\end{minipage} \\
\midrule
\endhead
Input \& data & Unlabelled corpus & sampled to 250,000 graphs & Fixed
compute cap per run. \\
& Labelled proxy & ESOL (regression), 5,000 labelled molecules & Proxy
only; used for selection. \\
& Unlabelled sampling cap & 250,000 sampled; 250,000 max graphs/run &
Constrains Phase-2 compute while keeping the objective comparison
meaningful. \\
& Graph caching & cache-datasets=1 & Keeps data pipeline constant across
trials. \\
Architecture & Backbone (gnn\_type) & GINE, D-MPNN, SchNet3D & Backbone
is part of the selection target. \\
& Hidden width & \{256, 384\} & Larger width is tested to check capacity
sensitivity. \\
& Depth & \{3, 4\} layers & Held small to keep compute bounded and
comparisons clean. \\
& 3D features & add\_3d ∈ \{0,1\} & Tests whether explicitly injecting
geometry helps the proxy under JEPA. \\
Masking \& augmentation & Contiguity & contiguity ∈ \{0,1\} & Toggles
``connected fragment'' masking vs non-contiguous masking. \\
& Mask ratio & 0.100--0.215 & Difficulty of the context→target
prediction task. \\
& Augmentation regime & augmentation\_profile ∈ \{none, graph\_light,
geom\_only, mixed\_light\} & Bundles chemically motivated perturbations
into regimes instead of toggling each operator independently. \\
Training dynamics & Learning rate & 2.1e-5 → 4.26e-4 & Broad log-range
to expose stability bands. \\
& EMA decay & 0.951 → 0.990 & Teacher update rate for JEPA stability. \\
& Pretrain budget & 2 epochs, max\_pretrain\_batches=800 & Short,
consistent compute budget per trial. \\
& Probe budget & 25 fine-tune epochs & Longer probing improves ranking
reliability at small label counts. \\
& Batch sizes & pretrain 256, finetune 256 & Held constant once the
Phase-2 protocol is set. \\
Runtime & Devices & 1 GPU & Single-GPU trials for consistent timing. \\
& Mixed precision & bf16=1 & Fixed across trials. \\
\bottomrule
\end{longtable}\par}

\begin{sidewaystable*}[tp]
\centering
\footnotesize
\renewcommand{\arraystretch}{1.3}\setlength{\tabcolsep}{4pt}
\caption*{\textbf{Table S5.} ESOL proxy RMSE by augmentation profile (Phase-2 sweep).}
\resizebox{0.95\textheight}{!}{%
\begin{tabular}{@{}
  >{\raggedright\arraybackslash}p{0.1070\linewidth}
  >{\raggedright\arraybackslash}p{0.0697\linewidth}
  >{\raggedright\arraybackslash}p{0.0655\linewidth}
  >{\raggedright\arraybackslash}p{0.0873\linewidth}
  >{\raggedright\arraybackslash}p{0.0832\linewidth}
  >{\raggedright\arraybackslash}p{0.0615\linewidth}
  >{\raggedright\arraybackslash}p{0.0956\linewidth}
  >{\raggedright\arraybackslash}p{0.0606\linewidth}
  >{\raggedright\arraybackslash}p{0.0707\linewidth}
  >{\raggedright\arraybackslash}p{0.0779\linewidth}
  >{\raggedright\arraybackslash}p{0.0606\linewidth}
  >{\raggedright\arraybackslash}p{0.0606\linewidth}@{}}
\toprule
\begin{minipage}[b]{\linewidth}\raggedright
\textbf{Regime}
\end{minipage} & \begin{minipage}[b]{\linewidth}\raggedright
\textbf{Rotate}
\end{minipage} & \begin{minipage}[b]{\linewidth}\raggedright
\textbf{Bond-Angle}
\end{minipage} & \begin{minipage}[b]{\linewidth}\raggedright
\textbf{Dihedral}
\end{minipage} & \begin{minipage}[b]{\linewidth}\raggedright
\textbf{Bond Deletion}
\end{minipage} & \begin{minipage}[b]{\linewidth}\raggedright
\textbf{Atom Mask}
\end{minipage} & \begin{minipage}[b]{\linewidth}\raggedright
\textbf{Subgraph Remove}
\end{minipage} & \begin{minipage}[b]{\linewidth}\raggedright
\textbf{n}
\end{minipage} & \begin{minipage}[b]{\linewidth}\raggedright
\textbf{Mean RMSE}
\end{minipage} & \begin{minipage}[b]{\linewidth}\raggedright
\textbf{Median RMSE}
\end{minipage} & \begin{minipage}[b]{\linewidth}\raggedright
\textbf{Min}
\end{minipage} & \begin{minipage}[b]{\linewidth}\raggedright
\textbf{Max}
\end{minipage} \\
\midrule\addlinespace[2pt]

none & ✗ & ✗ & ✗ & ✗ & ✗ & ✗ & 7 & 1.9714 & 1.984 & 1.698 & 2.241 \\
graph\_light & ✗ & ✗ & ✗ & ✓ * & ✓ * & ✗ & 40 & 2.046 & 2.048 & 1.696 &
2.490 \\
geom\_only & ✓ & ✓ & ✓ & ✗ & ✗ & ✗ & 16 & 3.711 & 3.731 & 3.353 &
4.153 \\
mixed\_light & ✓ & ✓ & ✓ & ✓ * & ✓ * & ✗ & 17 & 3.594 & 3.529 & 3.213 &
3.875 \\
\bottomrule
\end{tabular}
}
\end{sidewaystable*}

\subsection{S2.3. Statistical significance testing (Phase 1 paired comparison)}

Across the 12 matched backbone×seed pairs in Table S3, the contrastive
baseline exhibits higher ESOL proxy RMSE than M‑JEPA in every pair (mean
ΔRMSE = 1.6107). A nonparametric Wilcoxon signed‑rank test (one‑sided,
ΔRMSE \textgreater{} 0) rejects the null with p = 2.44e-04. A bootstrap
95\% confidence interval for the mean paired difference is {[}1.5181,
1.7169{]} RMSE.

\textbf{S2.4. Ablation scope and interpretation}

Phase 1 fixes connected masking (contiguity = 1) as part of the JEPA
view definition while varying other sweep dimensions for
objective-family screening under matched compute. Phase 2 co-varies
augmentation, masking (mask ratio and contiguity), and optimization
settings to identify a stable downstream configuration; these sweeps
provide sensitivity evidence but do not constitute controlled ablations
of any single component. In particular, we did not evaluate fixed-target
or random-target teacher variants, so the EMA teacher is treated as part
of the JEPA design rather than an isolated causal factor in this study.

\section{S3. SOTA Context Tables}

Published Tox21 results are included for contextual reference only.
Direct, fully matched comparison is generally not feasible because many
SOTA reports use random or cross-validation splits, multi-task training,
and optimize aggregate ROC-AUC, whereas this work emphasizes strict
scaffold splits, per-assay single-task evaluation, and reliability
reporting. (Wu et al., 2017; Mayr et al., 2016; Wang et al., 2021; Born
et al., 2023).

\subsection{S3.1 Aggregate SOTA Summary}

{\centering
\footnotesize
\renewcommand{\arraystretch}{1.3}\setlength{\tabcolsep}{3pt}
\begin{longtable}{@{}>{\raggedright\arraybackslash}p{0.1600\linewidth}>{\raggedright\arraybackslash}p{0.1640\linewidth}>{\raggedright\arraybackslash}p{0.1640\linewidth}>{\raggedright\arraybackslash}p{0.1640\linewidth}>{\raggedright\arraybackslash}p{0.1640\linewidth}>{\raggedright\arraybackslash}p{0.1640\linewidth}@{}}
\caption*{\textbf{Table S6.} Literature-reported Tox21 aggregate ROC-AUC results (heterogeneous protocols; provided for contextual reference). Note: many entries are multi-task and optimise/report only mean ROC-AUC aggregates under varying preprocessing and split regimes (random, cross-validation, scaffold). Per-assay outputs and calibration metrics are rarely reported, limiting direct comparison to our single-task scaffold-split evaluation.}\\
\toprule
\begin{minipage}[b]{\linewidth}\raggedright
\textbf{Model}
\end{minipage} & \begin{minipage}[b]{\linewidth}\raggedright
\textbf{Data}
\end{minipage} & \begin{minipage}[b]{\linewidth}\raggedright
\textbf{Work}
\end{minipage} & \begin{minipage}[b]{\linewidth}\raggedright
\textbf{ROC-AUC}
\end{minipage} & \begin{minipage}[b]{\linewidth}\raggedright
\textbf{Split regime}
\end{minipage} & \begin{minipage}[b]{\linewidth}\raggedright
\textbf{Task type}
\end{minipage} \\
\midrule
\endfirsthead
\toprule
\begin{minipage}[b]{\linewidth}\raggedright
\textbf{Model}
\end{minipage} & \begin{minipage}[b]{\linewidth}\raggedright
\textbf{Data}
\end{minipage} & \begin{minipage}[b]{\linewidth}\raggedright
\textbf{Work}
\end{minipage} & \begin{minipage}[b]{\linewidth}\raggedright
\textbf{ROC-AUC}
\end{minipage} & \begin{minipage}[b]{\linewidth}\raggedright
\textbf{Split regime}
\end{minipage} & \begin{minipage}[b]{\linewidth}\raggedright
\textbf{Task type}
\end{minipage} \\
\midrule
\endhead
SCFP (CNN) & T21C & (Hirohara et al., 2018) & 0.877 & Cross validation &
Multi-task * \\
DeepTox (DNN) & T21C & (Mayr et al., 2016) & 0.754 ± 0.144 & Cross
validation & Multi-task * \\
DeepTox (DNN) & T21C & (Mayr et al., 2016) & 0.723 ± 0.136 & Cross
validation & Single task * \\
Weave & T21C & (Kearnes et al., 2016) & 0.88 ± 0.003 & Cross validation
& Multi-task \\
FP2Vec & MN & (Jeon \& Kim, 2019) & 0.876 ± 0.006 & Random &
Multi-task \\
FP2Vec & MN & (Jeon \& Kim, 2019) & 0.825 ± 0.011 & Random & Single
task \\
Graph Conv. & MN & (Wu et al., 2017) & 0.829 ± 0.01 & Random &
Multi-task \\
HRGCN+ & MN & (Wu et al., 2021) & 0.848 ± 0.014 & Random & Multi-task \\
Attentive FP & MN & (Xiong et al., 2020) & 0.858 & Random &
Multi-task \\
D-MPNN & MN & (Yang et al., 2019) & 0.851 ± 0.002 & Random &
Multi-task \\
SELU-MPNN & MN & (Withnall et al., 2020) & 0.820 ± 0.01 & Random &
Multi-task \\
AMPNN & MN & (Withnall et al., 2020) & 0.812 ± 0.02 & Random &
Multi-task \\
EMPNN & MN & (Withnall et al., 2020) & 0.829 ± 0.01 & Random &
Multi-task \\
MMNB (MolMapNet) & MN & (Shen et al., 2021) & 0.842 & Random &
Multi-task \\
CLM & T21C & (Born et al., 2023) & 0.858 ± 0.05 & Random & Multi-task \\
Mol-BERT & MN & (Li \& Jiang, 2021) & 0.839 ± 0.075 & Scaffold &
Multi-task \\
MPNN & MN & (Li \& Jiang, 2021) & 0.808 ± 0.024 & Scaffold &
Multi-task \\
SMILES-BERT & MN & (Li \& Jiang, 2021) & 0.732 ± 0.025 & Scaffold &
Multi-task \\
GCN & MN & (Wang et al., 2021) & 70.9\% ± 2.6 & Random & Multi-task \\
GIN & MN & (Wang et al., 2021) & 74.0\% ± 0.8 & Random & Multi-task \\
SchNet & MN & (Wang et al., 2021) & 77.2\% ± 2.3 & Random &
Multi-task \\
MGCN & MN & (Wang et al., 2021) & 70.7\% ± 1.6 & Random & Multi-task \\
D-MPNN & MN & (Wang et al., 2021) & 68.9\% ± 1.3 & Random &
Multi-task \\
SchNet~ & MN & (Wang et al., 2021) & 77.2\% ± 2.3 & Random &
Multi-task \\
MolCLR (GCN) & MN & (Wang et al., 2021) & 74.7\% ± 0.8 & Random &
Multi-task \\
MolCLR (GIN) & MN & (Wang et al., 2021) & 79.8\% ± 0.7 & Random &
Multi-task \\
GIN (* Fine tune) & MN & (Hu, Liu, et al., 2020) & 76.7 \% ± 1.4 (avg) &
Scaffold & Multi-task \\
HiPM & MN & (Kang et al., 2024) & 0.770 ± 0.005 & Scaffold &
Multi-task \\
HiPM & MN & (Kang et al., 2024) & 0.799 ± 0.007 & Scaffold & Task
aware \\
HiPM & MN & (Kang et al., 2024) & 0.843 ± 0.007 & Scaffold &
Multi-task \\
GraphSAGE & MN & (Zang et al., 2023) & 69.9\% ± 1.1 & Scaffold &
Multi-task \\
GPT\_GNN & MN & (Zang et al., 2023) & 73.1\% ± 0.7 & Scaffold &
Multi-task \\
AttributeMask & MN & (Zang et al., 2023) & 74.6\% ± 0.3 & Scaffold &
Multi-task \\
ContextPred & MN & (Zang et al., 2023) & 72.7\% ± 0.6 & Scaffold &
Multi-task \\
InfoGraph & MN & (Zang et al., 2023) & 74.7\% ± 0.4 & Scaffold &
Multi-task \\
MoCL & MN & (Zang et al., 2023) & 70.9\% ± 0.2 & Scaffold &
Multi-task \\
GraphLoG & MN & (Zang et al., 2023) & 73.4\% ± 0.3 & Scaffold &
Multi-task \\
JOAO & MN & (Zang et al., 2023) & 75.5\% ± 0.7 & Scaffold &
Multi-task \\
G\_Motif & MN & (Zang et al., 2023) & 73.3\% ± 0.8 & Scaffold &
Multi-task \\
MGSSL & MN & (Zang et al., 2023) & 76.5\% ± 0.3 & Scaffold &
Multi-task \\
HiMol (large) & MN & (Zang et al., 2023) & 76.2\% ± 0.3 & Scaffold &
Multi-task \\
ImageMol & T21C & (Zeng et al., 2022) & 0.847 & Scaffold & Multi-task \\
SMO & T21C & (Idakwo et al., 2020) & 0.8069 & Random & Multi-task * \\
SMN & T21C & (Idakwo et al., 2020) & 0.8384 & Random & Multi-task * \\
Grover (large) & MN & (Rong et al., 2020) & 0.831 ± 0.025 & Scaffold &
Multi-task \\
OGB (GCN) & MN & (Hu, Fey, et al., 2020) & 86.03\% ± 1.37 & Random &
Multi-task \\
OGB (GIN) & MN & (Hu, Fey, et al., 2020) & 77.57\% ± 0.62 & Random &
Multi-task \\
SA-MTL & T21C & (Lim \& Lee, 2021) & 0.9 & Random & Multi-task \\
\bottomrule
\end{longtable}\par}

MN: MolculeNet data set,

T21C: Tox21 Challenge data set,

* indicates availability of metrics at per assay level covered below

\subsection{S3.2 Assay-Specific SOTA Metrics (where available)}

\begin{sidewaystable*}[tp]
\centering
\footnotesize
\renewcommand{\arraystretch}{1.3}\setlength{\tabcolsep}{4pt}
\caption*{\textbf{Table S7.} Assay-specific metrics compiled from prior work (where reported). Note: per-assay results are rarely available for scaffold-split settings; many sources use random or cross-validation splits and multi-task training, and calibration metrics are generally not reported. These values are included to contextualize typical ranges, not to provide protocol-matched baselines.}
\resizebox{0.95\textheight}{!}{%
\begin{tabular}{@{}
  >{\raggedright\arraybackslash}p{0.0986\linewidth}
  >{\raggedright\arraybackslash}p{0.0892\linewidth}
  >{\raggedright\arraybackslash}p{0.0893\linewidth}
  >{\raggedright\arraybackslash}p{0.0668\linewidth}
  >{\raggedright\arraybackslash}p{0.0809\linewidth}
  >{\raggedright\arraybackslash}p{0.0809\linewidth}
  >{\raggedright\arraybackslash}p{0.0668\linewidth}
  >{\raggedright\arraybackslash}p{0.0829\linewidth}
  >{\raggedright\arraybackslash}p{0.0829\linewidth}
  >{\raggedright\arraybackslash}p{0.0809\linewidth}
  >{\raggedright\arraybackslash}p{0.0809\linewidth}@{}}
\toprule
\begin{minipage}[b]{\linewidth}\raggedright
\textbf{Assay}
\end{minipage} & \begin{minipage}[b]{\linewidth}\raggedright
\textbf{AUC DeepTox}
\end{minipage} & \begin{minipage}[b]{\linewidth}\raggedright
\textbf{AUC DeepTox}
\end{minipage} & \begin{minipage}[b]{\linewidth}\raggedright
\textbf{AUC SCFP}
\end{minipage} & \begin{minipage}[b]{\linewidth}\raggedright
\textbf{AUC SMO}
\end{minipage} & \begin{minipage}[b]{\linewidth}\raggedright
\textbf{AUC SMN}
\end{minipage} & \begin{minipage}[b]{\linewidth}\raggedright
\textbf{AUC Mean}
\end{minipage} & \begin{minipage}[b]{\linewidth}\raggedright
\textbf{AUPRC SMO}
\end{minipage} & \begin{minipage}[b]{\linewidth}\raggedright
\textbf{AUPRC SMN}
\end{minipage} & \begin{minipage}[b]{\linewidth}\raggedright
\textbf{Brier SMO}
\end{minipage} & \begin{minipage}[b]{\linewidth}\raggedright
\textbf{Brier SMN}
\end{minipage} \\
\midrule\addlinespace[2pt]

Data & T21C & T21C & T21C & T21C & T21C & & T21C & T21C & T21C & T21C \\
Split & Cros Val & Cros Val & Cros Val & Random & Random & & Random &
Random & Random & Random \\
Learning & Multi & Single & Multi & Multi & Multi & & Multi & Multi &
Multi & Multi \\
NR-AhR & 0.8409 & 0.8487 & 0.921 & 0.8936 & 0.9196 & 0.885 ± 0.034 &
0.5065 & 0.5660 & 0.3281 & 0.2583 \\
NR-AR & 0.3459 & 0.3755 & 0.864 & 0.7780 & 0.6810 & 0.609 ± 0.211 &
0.3290 & 0.0685 & 0.4263 & 0.4303 \\
NR-AR-LBD & 0.9289 & 0.8799 & 0.908 & 0.7509 & 0.7969 & 0.853 ± 0.068 &
0.0821 & 0.0639 & 0.6739 & 0.4156 \\
NR-Aromatase & 0.7921 & 0.7523 & 0.861 & 0.8112 & 0.8500 & 0.813 ± 0.039
& 0.3504 & 0.3845 & 0.3379 & 0.3327 \\
NR-ER & 0.6949 & 0.6659 & 0.792 & 0.7296 & 0.8628 & 0.749 ± 0.071 &
0.3895 & 0.5688 & 0.4205 & 0.3134 \\
NR-ER-LBD & 0.7272 & 0.6532 & 0.884 & 0.8072 & 0.8233 & 0.779 ± 0.080 &
0.2658 & 0.2018 & 0.4067 & 0.3670 \\
NR-PPAR-γ & 0.7102 & 0.6367 & 0.904 & 0.7872 & 0.7713 & 0.762 ± 0.089 &
0.2806 & 0.3736 & 0.4138 & 0.3503 \\
SR-ARE & 0.8017 & 0.7927 & 0.839 & 0.7714 & 0.8910 & 0.819 ± 0.042 &
0.4052 & 0.6443 & 0.3881 & 0.2761 \\
SR-ATAD5 & 0.7958 & 0.7972 & 0.929 & 0.8151 & 0.8093 & 0.829 ± 0.050 &
0.3350 & 0.2422 & 0.3924 & 0.3431 \\
SR-HSE & 0.8101 & 0.7354 & 0.826 & 0.7983 & 0.8483 & 0.804 ± 0.038 &
0.1993 & 0.1134 & 0.4146 & 0.3491 \\
SR-MMP & 0.8489 & 0.8485 & 0.914 & 0.8893 & 0.9294 & 0.886 ± 0.033 &
0.4928 & 0.5234 & 0.3467 & 0.2371 \\
SR-p53 & 0.7487 & 0.6955 & 0.907 & 0.8510 & 0.8785 & 0.816 ± 0.081 &
0.2913 & 0.3254 & 0.3814 & 0.3014 \\
Mean ± (column) & 0.754 ± 0.138 & 0.723 ± 0.131 & 0.879 ± 0.041 & 0.807
± 0.048 & 0.838 ± 0.066 & 0.800 ± 0.071 & 0.327 ± 0.113 & 0.340 ± 0.196
& 0.411 ± 0.085 & 0.331 ± 0.056 \\
\bottomrule
\end{tabular}
}
\end{sidewaystable*}

\section{S4. Graph Construction Validation Probes}

To validate that generated graph objects are chemically faithful and
structurally consistent with SMILES, we performed visual probes and
graph-theoretic consistency checks. These checks are model-agnostic and
are intended to confirm that atoms map to nodes and covalent bonds map
to edges as expected, ring topology is preserved, and explicit hydrogen
handling yields chemically plausible degrees.

\begin{center}
\includegraphics[width=0.62\linewidth]{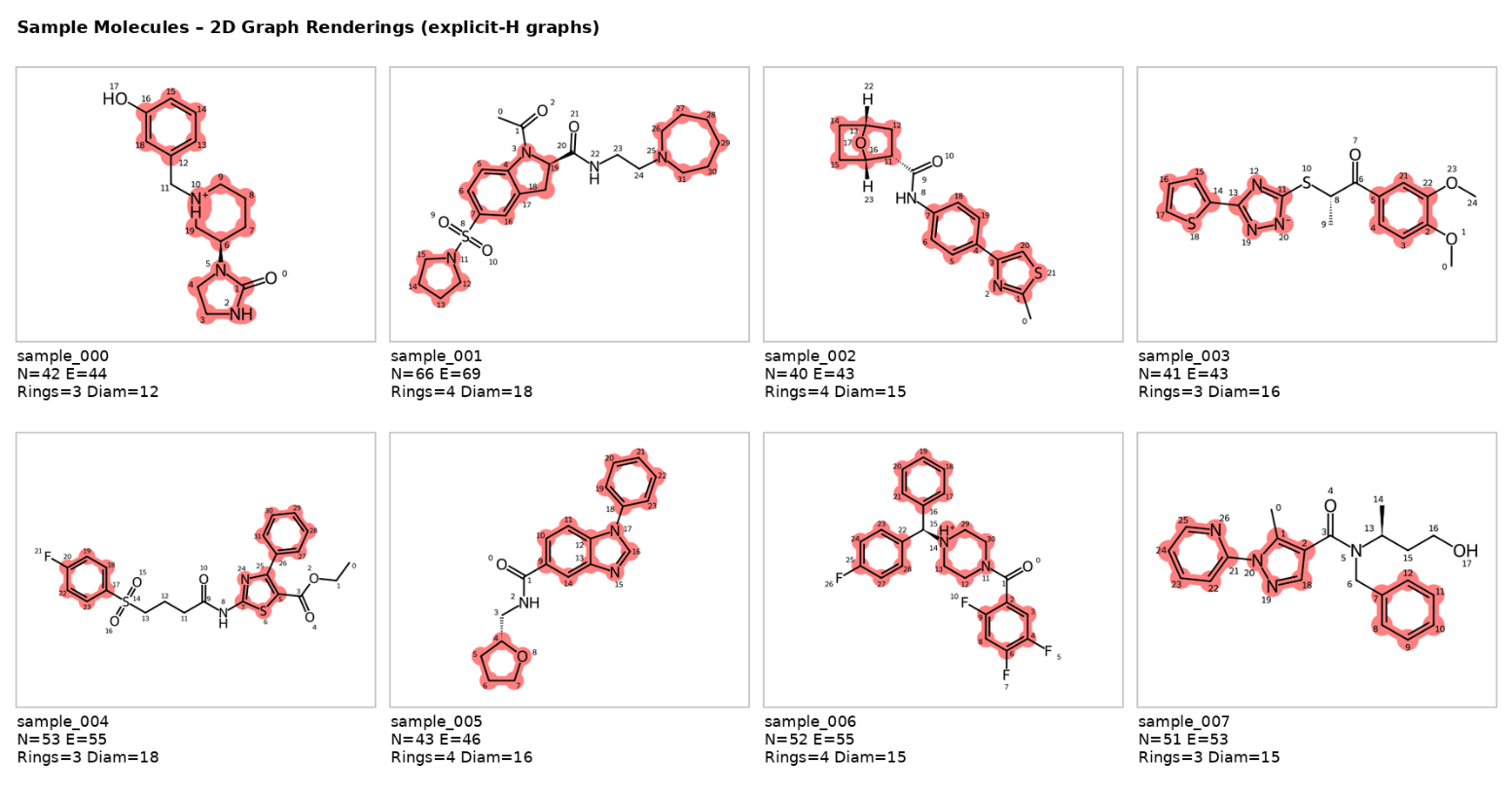}\\[2pt]
{\footnotesize\textbf{Figure S1.} Representative molecules rendered as explicit-hydrogen graphs for qualitative verification of connectivity and ring structure.}
\end{center}

\begin{center}
\includegraphics[width=0.42\linewidth]{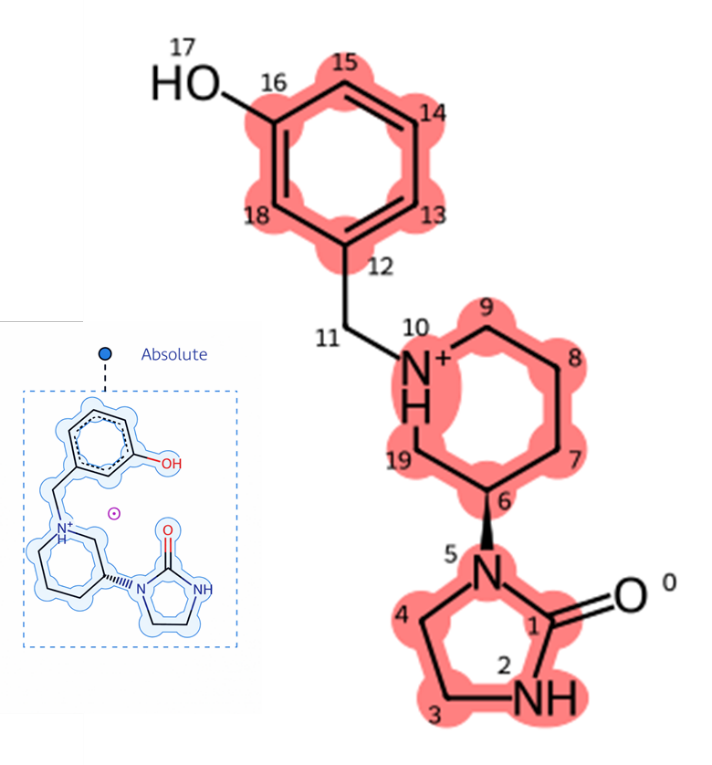}\\[2pt]
{\footnotesize\textbf{Figure S2.} Worked validation example. Canonical 2D structure and graph-annotated representation for Sample\_000 (SMILES: O=C1NCCN1{[}C@@H{]}1CCC{[}NH+{]}(Cc2cccc(O)c2)C1). For this molecule, \textbar V\textbar=42, \textbar E\textbar=44, mean degree=2\textbar E\textbar/\textbar V\textbar≈2.10, and cyclomatic number \textbar E\textbar−\textbar V\textbar+1=3, matching the three-ring scaffold.}
\end{center}

\textbf{S4.3 Augmentation validity scope.}

During dataset ingestion (ZINC, ESOL, and Tox21), molecules that fail
RDKit sanitization---including kekulization and aromaticity perception
failures-are removed prior to graph caching, ensuring that all base
graphs used for training and evaluation are chemically valid. These
sanitization checks are applied only at load time; we did not re-run
RDKit sanitization after applying graph augmentations during training,
nor did we log the fraction of molecules filtered during ingestion.
Consequently, we do not quantify post-augmentation chemical validity per
augmentation regime. The performance degradation observed under heavy
augmentation is therefore interpreted as an optimization-level
sensitivity under the fixed training budget, rather than as evidence of
chemically invalid augmented structures.

\section{S5. Tox21 Hybrid Evaluation Configuration}

Appendix D in the source thesis contained a placeholder. Here, we
provide the full downstream configuration used for the hybrid
scaffold-split Tox21 evaluation, including encoder/head learning rates,
early freezing, staged unfreezing, checkpoint selection metric, and
calibration-related parameters.

\subsection{S5.1 Global Hybrid Configuration}

{\centering
\small
\renewcommand{\arraystretch}{1.3}\setlength{\tabcolsep}{3pt}
\begin{longtable}{@{}>{\raggedright\arraybackslash}p{0.1600\linewidth}>{\raggedright\arraybackslash}p{0.2733\linewidth}>{\raggedright\arraybackslash}p{0.2733\linewidth}>{\raggedright\arraybackslash}p{0.2733\linewidth}@{}}
\caption*{\textbf{Table S8.} Downstream Tox21 hybrid evaluation configuration and key CLI/config parameters.}\\
\toprule
\begin{minipage}[b]{\linewidth}\raggedright
\textbf{Group}
\end{minipage} & \begin{minipage}[b]{\linewidth}\raggedright
\textbf{Parameter (CLI / config name)}
\end{minipage} & \begin{minipage}[b]{\linewidth}\raggedright
\textbf{Value(s)}
\end{minipage} & \begin{minipage}[b]{\linewidth}\raggedright
\textbf{what it controls}
\end{minipage} \\
\midrule
\endfirsthead
\toprule
\begin{minipage}[b]{\linewidth}\raggedright
\textbf{Group}
\end{minipage} & \begin{minipage}[b]{\linewidth}\raggedright
\textbf{Parameter (CLI / config name)}
\end{minipage} & \begin{minipage}[b]{\linewidth}\raggedright
\textbf{Value(s)}
\end{minipage} & \begin{minipage}[b]{\linewidth}\raggedright
\textbf{what it controls}
\end{minipage} \\
\midrule
\endhead
Input \& data & Dataset & Tox21 & 12 binary toxicity assays; strong
class imbalance makes PR-AUC and calibration metrics informative. \\
& -\/-csv & /srv/mjepa/data/tox21/data.csv & Source table for
labels/SMILES; used to generate graph batches. \\
& -\/-tasks & all -- not shown for brevity & Multi-assay evaluation:
per-assay metrics reported. \\
& Split strategy & Scaffold (implied by pipeline) & Scaffold splitting
reduces chemical leakage; typically lowers AUC vs random split. \\
Representation / encoder & -\/-encoder-checkpoint &
/data/mjepa/cache/pretrain/encoder.pt & Starting molecular encoder
weights used for hybrid fine-tuning. \\
& -\/-encoder-source / -\/-evaluation-mode & hybrid & Enables staged
hybrid schedule (freeze → unfreeze) and hybrid evaluation
bookkeeping. \\
\begin{minipage}[t]{\linewidth}\raggedright
Architecture\\
(from phase 2 sweep)\strut
\end{minipage} & -\/-gnn-type & D-MPNN & Message-passing graph encoder;
strong baseline for molecular property prediction. \\
& -\/-num-layers & 4 & Depth of message passing; controls receptive
field / expressivity. \\
& -\/-hidden-dim & 256 & Representation width; trade-off between
capacity and overfitting. \\
& -\/-dropout & 0.2 & Regularisation to improve generalisation on
small/imperfectly balanced assays. \\
& -\/-add-3d & 0 & 2D graph only; avoids confounding with conformer
quality/availability. \\
Optimisation (global) & -\/-lr & 1e-3 & Global learning rate used by
training script defaults. \\
& -\/-head-lr & 3e-4 & Step size for classification head; strongly
influences PR-AUC under imbalance. \\
& -\/-encoder-lr & 1e-5 & Conservative encoder update; stabilises
representation while allowing task adaptation. \\
& -\/-weight-decay & 1e-4 & L2 regularisation; reduces overconfident
probabilities that inflate ECE/Brier. \\
& -\/-lr-scheduler & cosine & Smooth annealing; helps avoid late-epoch
overfitting. \\
& -\/-warmup-ratio & 0.10 & Early stabilisation for mixed precision +
multi-GPU training. \\
& -\/-min-lr & 1e-6 & Floor for cosine schedule; prevents premature
convergence. \\
& -\/-layerwise-decay & 0.8 & Smaller updates in lower layers; preserves
learned chemistry priors while adapting top layers. \\
Hybrid schedule & -\/-hybrid-freeze-epochs & 15 & Freeze encoder early
to fit a stable head before representation drift. \\
& -\/-unfreeze-top-layers & 2 & Only top layers are unfrozen first;
balances plasticity vs stability. \\
& -\/-full-finetune & enabled & After warm start, allows end-to-end
adaptation for harder assays. \\
& -\/-epochs / -\/-patience & 100 / 12 & Long horizon with early
stopping; mitigates overtraining. \\
Ensembling / heads & -\/-head-ensemble-size & 3 & Number of linear
heads; supports mean±std reporting across heads (``seeds''). \\
& -\/-seeds & 0 1 2 3 4 & Independent training initialisations; used for
uncertainty/variance estimates. \\
& -\/-checkpoint-metric & PR-AUC & Chooses the checkpoint that optimises
minority-class retrieval (more appropriate than ROC-AUC under
imbalance). \\
Class imbalance controls & -\/-class-weights & auto & Sets per-assay
positive class weight from label frequency (pos\_weight ≈ neg/pos). \\
& -\/-triage-pct & 0.0 & Disables triage; uses full training set per
assay. \\
Calibration \& probability quality & -\/-temperature & 0.1 &
Calibrator/temperature parameter passed; affects probability scaling and
thus ECE/Brier. \\
& Calibration policy & (from pipeline) & If you're using per-head
calibration, ECE/Brier become directly comparable across heads and
ensemble. \\
Explainability & -\/-explain-mode & ig, ig\_motif & Integrated gradients
+ motif aggregation; supports chemical interpretability
(JCIM-friendly). \\
& -\/-explain-steps & No cap & IG integration steps; trade-off between
stability and compute. \\
Compute / reproducibility & -\/-devices / -\/-nproc\_per\_node & 4 &
Distributed training across 4 GPUs. \\
& -\/-bf16 & 1 & Mixed precision; improves throughput while preserving
stability in modern GPUs. \\
& -\/-batch-size & 256 & Effective batch size; affects gradient noise
and calibration. \\
& Data loader & \begin{minipage}[t]{\linewidth}\raggedright
-\/-num-workers 16,\\
-\/-prefetch-factor 4,\\
-\/-persistent-workers 1,\\
-\/-pin-memory 1\strut
\end{minipage} & Performance settings: should not change metrics but
affects runtime. \\
& -\/-verify-match-threshold & 0.98 & Encoder weight compatibility check
ensures checkpoint matches architecture. \\
\bottomrule
\end{longtable}\par}

\subsection{S5.2 Per-Assay Overrides}

{\centering
\small
\renewcommand{\arraystretch}{1.3}\setlength{\tabcolsep}{3pt}
\begin{longtable}{@{}>{\raggedright\arraybackslash}p{0.1600\linewidth}>{\raggedright\arraybackslash}p{0.2733\linewidth}>{\raggedright\arraybackslash}p{0.2733\linewidth}>{\raggedright\arraybackslash}p{0.2733\linewidth}@{}}
\caption*{\textbf{Table S9.} Assay-specific overridden parameters used to stabilize training under severe imbalance or noise.}\\
\toprule
\begin{minipage}[b]{\linewidth}\raggedright
\textbf{Assay}
\end{minipage} & \begin{minipage}[b]{\linewidth}\raggedright
\textbf{Overridden parameter(s)}
\end{minipage} & \begin{minipage}[b]{\linewidth}\raggedright
\textbf{Value(s)}
\end{minipage} & \begin{minipage}[b]{\linewidth}\raggedright
\textbf{Rationale}
\end{minipage} \\
\midrule
\endfirsthead
\toprule
\begin{minipage}[b]{\linewidth}\raggedright
\textbf{Assay}
\end{minipage} & \begin{minipage}[b]{\linewidth}\raggedright
\textbf{Overridden parameter(s)}
\end{minipage} & \begin{minipage}[b]{\linewidth}\raggedright
\textbf{Value(s)}
\end{minipage} & \begin{minipage}[b]{\linewidth}\raggedright
\textbf{Rationale}
\end{minipage} \\
\midrule
\endhead
NR-Aromatase & pos\_weight & 10 & Strong imbalance: explicit weighting
stabilises minority gradients. \\
NR-AR & pos\_weight, focal\_gamma & 8, 1.5 & Moderate imbalance; reduced
γ avoids over-penalising easy negatives. \\
SR-p53 & pos\_weight, focal\_gamma & 10, 1.5 & Noisy assay: softer focal
loss improves calibration stability. \\
SR-MMP & pos\_weight & 5.33 & Higher positive fraction; weaker weighting
sufficient. \\
\bottomrule
\end{longtable}\par}

\section{S6. Representation-only Transfer Benchmark}

To separate representation quality from task-specific optimization, the
selected M-JEPA encoder was evaluated using a representation-only
transfer benchmark in which fresh heads were trained on an external
labelled dataset across multiple random seeds. All Tox21-specific tuning
was removed. Results are reported for completeness and are not used to
support claims in the main manuscript.

{\centering
\small
\renewcommand{\arraystretch}{1.3}\setlength{\tabcolsep}{3pt}
\begin{longtable}{@{}>{\raggedright\arraybackslash}p{0.1600\linewidth}>{\raggedright\arraybackslash}p{0.4100\linewidth}>{\raggedright\arraybackslash}p{0.4100\linewidth}@{}}
\caption*{\textbf{Table S10.} Representation-only transfer benchmark summary.}\\
\toprule
\begin{minipage}[b]{\linewidth}\raggedright
\textbf{Metric}
\end{minipage} & \begin{minipage}[b]{\linewidth}\raggedright
\textbf{Mean ± SD}
\end{minipage} & \begin{minipage}[b]{\linewidth}\raggedright
\textbf{Notes}
\end{minipage} \\
\midrule
\endfirsthead
\toprule
\begin{minipage}[b]{\linewidth}\raggedright
\textbf{Metric}
\end{minipage} & \begin{minipage}[b]{\linewidth}\raggedright
\textbf{Mean ± SD}
\end{minipage} & \begin{minipage}[b]{\linewidth}\raggedright
\textbf{Notes}
\end{minipage} \\
\midrule
\endhead
RMSE & 1.63 ± 0.04 & Regression error on external labelled dataset;
fresh heads trained without Tox21-specific optimisation. \\
MAE & 1.26 ± 0.03 & Consistent with RMSE magnitude; reported for
completeness. \\
R² & 0.37 ± 0.03 & Moderate explanatory power; not directly comparable
to ESOL proxy RMSE due to dataset and protocol differences. \\
Validation RMSE & 2.01 ± 0.03 & Validation split performance; not used
for model selection. \\
Number of seeds & 3 & Independent head initialisations. \\
Train / Val / Test size & 901 / 113 / 113 & Explicit split strategy. \\
\bottomrule
\end{longtable}\par}

\section{S7. Attribution protocol and stability analysis}

We compute Integrated Gradients (IG) attributions for each fine‑tuned
Tox21 head with respect to node and edge input features, using a
zeroed‑feature baseline and a fixed integration path in input space.
Per‑atom scores are aggregated to chemically meaningful motifs by
pooling over ring systems and Bemis--Murcko scaffold components,
enabling motif‑level summaries that are less sensitive to atom indexing.

To assess attribution stability during fine‑tuning beyond a small number
of illustrative cases, we compute Spearman rank correlation of
motif‑level IG rank order between an early checkpoint (pretrained
initialization) and the late checkpoint (post fine‑tuning), using motif
ranks induced by \textbar normalized motif IG\textbar{} scores. We
evaluate 62 molecule--assay pairs spanning all 12 Tox21 endpoints (4--6
molecules per assay) selected via a pre‑specified stratified protocol
covering high‑confidence TP/TN, high‑confidence error, and borderline
cases (Table S14). Across this set, attribution stability is typically
high (median ρ = 0.90; IQR 0.61--1.00) but not guaranteed: 12/62 pairs
show ρ \textless{} 0.50, with the lowest stability concentrated in
NR‑ER‑LBD (median ρ = −0.23; Table S14). Table S12 retains three
illustrative molecules used in the main text for qualitative reference.

The qualitative case studies in the main text are intended as
diagnostics under scaffold shift rather than definitive mechanistic
claims; they should be interpreted alongside discrimination and
calibration metrics.

\section{S8. Full per‑assay discrimination and calibration metrics (Tox21)}

\begin{sidewaystable*}[tp]
\centering
\footnotesize
\renewcommand{\arraystretch}{1.3}\setlength{\tabcolsep}{4pt}
\caption*{\textbf{Table S11.} Per‑assay metrics on Tox21 scaffold split for baseline vs. JEPA‑hybrid models (mean ± SD across head ensemble).}
\resizebox{0.95\textheight}{!}{%
\begin{tabular}{@{}
  >{\raggedright\arraybackslash}p{0.0845\linewidth}
  >{\raggedright\arraybackslash}p{0.0738\linewidth}
  >{\raggedright\arraybackslash}p{0.0560\linewidth}
  >{\raggedright\arraybackslash}p{0.0703\linewidth}
  >{\raggedright\arraybackslash}p{0.0602\linewidth}
  >{\raggedright\arraybackslash}p{0.0524\linewidth}
  >{\raggedright\arraybackslash}p{0.0703\linewidth}
  >{\raggedright\arraybackslash}p{0.0602\linewidth}
  >{\raggedright\arraybackslash}p{0.0524\linewidth}
  >{\raggedright\arraybackslash}p{0.0703\linewidth}
  >{\raggedright\arraybackslash}p{0.0602\linewidth}
  >{\raggedright\arraybackslash}p{0.0590\linewidth}
  >{\raggedright\arraybackslash}p{0.0703\linewidth}
  >{\raggedright\arraybackslash}p{0.0602\linewidth}@{}}
\toprule
\begin{minipage}[b]{\linewidth}\raggedright
\textbf{Assay}
\end{minipage} & \begin{minipage}[b]{\linewidth}\raggedright
\textbf{+ve Fraction}
\end{minipage} & \begin{minipage}[b]{\linewidth}\raggedright
\textbf{ROC-AUC\\
(Base ±σ)}\strut
\end{minipage} & \begin{minipage}[b]{\linewidth}\raggedright
\textbf{ROC-AUC\\
(Hybrid ±σ)}\strut
\end{minipage} & \begin{minipage}[b]{\linewidth}\raggedright
\textbf{Δ ROC}
\end{minipage} & \begin{minipage}[b]{\linewidth}\raggedright
\textbf{PR-AUC\\
(Base ±σ)}\strut
\end{minipage} & \begin{minipage}[b]{\linewidth}\raggedright
\textbf{PR-AUC\\
(Hybrid ±σ)}\strut
\end{minipage} & \begin{minipage}[b]{\linewidth}\raggedright
\textbf{Δ PR}
\end{minipage} & \begin{minipage}[b]{\linewidth}\raggedright
\textbf{Brier\\
(Base ±σ)}\strut
\end{minipage} & \begin{minipage}[b]{\linewidth}\raggedright
\textbf{Brier\\
(Hybrid ±σ)}\strut
\end{minipage} & \begin{minipage}[b]{\linewidth}\raggedright
\textbf{Δ Brier}
\end{minipage} & \begin{minipage}[b]{\linewidth}\raggedright
\textbf{ECE\\
(Base ±σ)}\strut
\end{minipage} & \begin{minipage}[b]{\linewidth}\raggedright
\textbf{ECE\\
(Hybrid ±σ)}\strut
\end{minipage} & \begin{minipage}[b]{\linewidth}\raggedright
\textbf{Δ ECE}
\end{minipage} \\
\midrule\addlinespace[2pt]

NR-AR & 0.0425 & 0.545 ± 0.085 & 0.625 ± 0.002 & +0.080 & 0.078 ± 0.037
& 0.118 ± 0.001 & +0.041 & 0.040 ± 0.007 & 0.040 ± 0.006 & −0.001 &
0.0149 ± 0.010 & 0.0107 ± 0.016 & −0.004 \\
NR-AR-LBD & 0.0351 & 0.678 ± 0.057 & 0.696 ± 0.057 & +0.018 & 0.081 ±
0.036 & 0.098 ± 0.036 & +0.017 & 0.031 ± 0.068 & 0.028 ± 0.091 & −0.001
& 0.0100 ± 0.082 & 0.0276 ± 0.091 & +0.018 \\
NR-AhR & 0.1173 & 0.626 ± 0.082 & 0.739 ± 0.008 & +0.113 & 0.212 ± 0.016
& 0.312 ± 0.016 & +0.100 & 0.258 ± 0.083 & 0.395 ± 0.078 & +0.137 &
0.363 ± 0.145 & 0.153 ± 0.078 & −0.210 \\
NR-Aromatase & 0.0515 & 0.477 ± 0.131 & 0.561 ± 0.004 & +0.084 & 0.103 ±
0.002 & 0.143 ± 0.002 & +0.039 & 0.253 ± 0.131 & 0.089 ± 0.011 & −0.164
& 0.406 ± 0.140 & 0.025 ± 0.017 & −0.381 \\
NR-ER & 0.1280 & 0.499 ± 0.096 & 0.569 ± 0.013 & +0.070 & 0.170 ± 0.012
& 0.189 ± 0.012 & +0.019 & 0.257 ± 0.096 & 0.252 ± 0.004 & −0.006 &
0.361 ± 0.221 & 0.220 ± 0.095 & −0.140 \\
NR-ER-LBD & 0.0503 & 0.534 ± 0.044 & 0.545 ± 0.014 & +0.011 & 0.069 ±
0.003 & 0.061 ± 0.012 & −0.008 & 0.180 ± 0.044 & 0.041 ± 0.012 & −0.138
& 0.372 ± 0.058 & 0.001 ± 0.001 & −0.370 \\
NR-PPAR-γ & 0.0288 & 0.427 ± 0.084 & 0.635 ± 0.013 & +0.208 & 0.053 ±
0.012 & 0.085 ± 0.012 & +0.033 & 0.057 ± 0.084 & 0.057 ± 0.022 &
\textasciitilde0.000 & 0.022 ± 0.084 & 0.022 ± 0.022 &
\textasciitilde0.000 \\
SR-ARE & 0.1615 & 0.526 ± 0.083 & 0.532 ± 0.004 & +0.006 & 0.240 ± 0.016
& 0.244 ± 0.016 & +0.003 & 0.254 ± 0.083 & 0.213 ± 0.017 & −0.009 &
0.275 ± 0.083 & 0.185 ± 0.018 & −0.191 \\
SR-ATAD5 & 0.0373 & 0.571 ± 0.142 & 0.674 ± 0.014 & +0.103 & 0.059 ±
0.012 & 0.082 ± 0.012 & +0.023 & 0.255 ± 0.142 & 0.052 ± 0.021 & −0.203
& 0.456 ± 0.142 & 0.021 ± 0.021 & −0.441 \\
SR-HSE & 0.0575 & 0.505 ± 0.133 & 0.556 ± 0.013 & +0.051 & 0.098 ± 0.012
& 0.122 ± 0.012 & +0.024 & 0.252 ± 0.133 & 0.077 ± 0.021 & −0.181 &
0.427 ± 0.133 & 0.077 ± 0.077 & −0.400 \\
SR-MMP & 0.1580 & 0.537 ± 0.086 & 0.554 ± 0.013 & +0.017 & 0.259 ± 0.016
& 0.272 ± 0.016 & +0.012 & 0.255 ± 0.086 & 0.237 ± 0.017 & −0.008 &
0.299 ± 0.086 & 0.266 ± 0.016 & −0.033 \\
SR-p53 & 0.0624 & 0.554 ± 0.051 & 0.553 ± 0.013 & −0.001 & 0.125 ± 0.009
& 0.134 ± 0.009 & +0.009 & 0.224 ± 0.051 & 0.163 ± 0.022 & −0.061 &
0.346 ± 0.051 & 0.097 ± 0.022 & −0.249 \\
\bottomrule
\end{tabular}
}
\end{sidewaystable*}

{\centering
\footnotesize
\renewcommand{\arraystretch}{1.3}\setlength{\tabcolsep}{3pt}
\begin{longtable}{@{}>{\raggedright\arraybackslash}p{0.1600\linewidth}>{\raggedright\arraybackslash}p{0.2050\linewidth}>{\raggedright\arraybackslash}p{0.2050\linewidth}>{\raggedright\arraybackslash}p{0.2050\linewidth}>{\raggedright\arraybackslash}p{0.2050\linewidth}@{}}
\caption*{\textbf{Table S12.} Attribution stability between early and late fine‑tuning checkpoints for three illustrative molecules. Spearman ρ is computed over motif‑rank order; ΔIG statistics report absolute change in normalized motif IG scores. Aggregate stability statistics across 62 molecule--assay pairs are reported in Table S14.}\\
\toprule
\begin{minipage}[b]{\linewidth}\raggedright
\textbf{Molecule}
\end{minipage} & \begin{minipage}[b]{\linewidth}\raggedright
\textbf{Spearman ρ}
\end{minipage} & \begin{minipage}[b]{\linewidth}\raggedright
\textbf{Max \textbar ΔIG\textbar{}}
\end{minipage} & \begin{minipage}[b]{\linewidth}\raggedright
\textbf{Mean \textbar ΔIG\textbar{}}
\end{minipage} & \begin{minipage}[b]{\linewidth}\raggedright
\textbf{Median \textbar ΔIG\textbar{}}
\end{minipage} \\
\midrule
\endfirsthead
\toprule
\begin{minipage}[b]{\linewidth}\raggedright
\textbf{Molecule}
\end{minipage} & \begin{minipage}[b]{\linewidth}\raggedright
\textbf{Spearman ρ}
\end{minipage} & \begin{minipage}[b]{\linewidth}\raggedright
\textbf{Max \textbar ΔIG\textbar{}}
\end{minipage} & \begin{minipage}[b]{\linewidth}\raggedright
\textbf{Mean \textbar ΔIG\textbar{}}
\end{minipage} & \begin{minipage}[b]{\linewidth}\raggedright
\textbf{Median \textbar ΔIG\textbar{}}
\end{minipage} \\
\midrule
\endhead
1 & 0.472 & 0.350 & 0.0564 & 0.0154 \\
2 & 0.946 & 0.125 & 0.0092 & 0.0030 \\
3 & 0.912 & 0.129 & 0.0167 & 0.0120 \\
\bottomrule
\end{longtable}\par}

\section{S9. Cross-assay significance testing (Tox21)}

To quantify whether base-to-hybrid deltas are systematically nonzero
across endpoints, we treat each of the 12 Tox21 assays as a paired
observation and report (i) a two-sided Wilcoxon signed-rank test for the
median delta and (ii) a nonparametric bootstrap 95\% confidence interval
for the mean delta across assays (10,000 resamples). Results for
discrimination (ROC-AUC, PR-AUC) and reliability (Brier, ECE) are
summarized in Table S13.

{\centering
\footnotesize
\renewcommand{\arraystretch}{1.3}\setlength{\tabcolsep}{3pt}
\begin{longtable}{@{}>{\raggedright\arraybackslash}p{0.1600\linewidth}>{\raggedright\arraybackslash}p{0.1640\linewidth}>{\raggedright\arraybackslash}p{0.1640\linewidth}>{\raggedright\arraybackslash}p{0.1640\linewidth}>{\raggedright\arraybackslash}p{0.1640\linewidth}>{\raggedright\arraybackslash}p{0.1640\linewidth}@{}}
\caption*{\textbf{Table S13.} Cross-assay significance tests for base-to-hybrid deltas on the Tox21 scaffold split (Δ = Hybrid − Base, computed per endpoint from the ensemble prediction). For ROC-AUC and PR-AUC, positive Δ indicates improvement; for Brier and ECE, negative Δ indicates improvement. Bootstrap confidence intervals are computed for the mean Δ across assays (10,000 resamples). Wilcoxon p-values are two-sided tests of whether the median Δ differs from zero.}\\
\toprule
\begin{minipage}[b]{\linewidth}\raggedright
\textbf{Metric}
\end{minipage} & \begin{minipage}[b]{\linewidth}\raggedright
\textbf{Mean Δ}
\end{minipage} & \begin{minipage}[b]{\linewidth}\raggedright
\textbf{Median Δ}
\end{minipage} & \begin{minipage}[b]{\linewidth}\raggedright
\textbf{Improved assays}
\end{minipage} & \begin{minipage}[b]{\linewidth}\raggedright
\textbf{95\% bootstrap CI (mean Δ)}
\end{minipage} & \begin{minipage}[b]{\linewidth}\raggedright
\textbf{Wilcoxon p (two-sided)}
\end{minipage} \\
\midrule
\endfirsthead
\toprule
\begin{minipage}[b]{\linewidth}\raggedright
\textbf{Metric}
\end{minipage} & \begin{minipage}[b]{\linewidth}\raggedright
\textbf{Mean Δ}
\end{minipage} & \begin{minipage}[b]{\linewidth}\raggedright
\textbf{Median Δ}
\end{minipage} & \begin{minipage}[b]{\linewidth}\raggedright
\textbf{Improved assays}
\end{minipage} & \begin{minipage}[b]{\linewidth}\raggedright
\textbf{95\% bootstrap CI (mean Δ)}
\end{minipage} & \begin{minipage}[b]{\linewidth}\raggedright
\textbf{Wilcoxon p (two-sided)}
\end{minipage} \\
\midrule
\endhead
ROC-AUC & +0.048 & +0.030 & 12/12 & {[}0.025, 0.078{]} &
4.88×10\^{}-4 \\
PR-AUC & +0.026 & +0.016 & 11/12 & {[}0.013, 0.042{]} & 9.77×10\^{}-4 \\
Brier & −0.068 & −0.029 & 12/12 & {[}−0.114, −0.028{]} &
4.88×10\^{}-4 \\
ECE & −0.216 & −0.229 & 10/12 & {[}−0.300, −0.129{]} & 4.88×10\^{}-3 \\
\bottomrule
\end{longtable}\par}

{\centering
\footnotesize
\renewcommand{\arraystretch}{1.3}\setlength{\tabcolsep}{3pt}
\begin{longtable}{@{}>{\raggedright\arraybackslash}p{0.1600\linewidth}>{\raggedright\arraybackslash}p{0.1640\linewidth}>{\raggedright\arraybackslash}p{0.1640\linewidth}>{\raggedright\arraybackslash}p{0.1640\linewidth}>{\raggedright\arraybackslash}p{0.1640\linewidth}>{\raggedright\arraybackslash}p{0.1640\linewidth}@{}}
\caption*{\textbf{Table S14.} Aggregate motif‑IG attribution stability across Tox21 endpoints. Spearman ρ is computed between early (pretrained) and late (fine‑tuned) checkpoints over motif rank order induced by \textbar normalized motif IG\textbar{} scores. Molecules are selected per endpoint via a stratified protocol (high‑confidence TP/TN, high‑confidence error, borderline). ``Frac. low'' reports the fraction of cases with ρ \textless{} 0.50. ``Median Top‑3 overlap'' is the median Jaccard overlap between the sets of top‑3 motifs by \textbar IG\textbar{} at early vs late checkpoints.}\\
\toprule
\begin{minipage}[b]{\linewidth}\raggedright
\textbf{Assay}
\end{minipage} & \begin{minipage}[b]{\linewidth}\raggedright
\textbf{n}
\end{minipage} & \begin{minipage}[b]{\linewidth}\raggedright
\textbf{Median ρ}
\end{minipage} & \begin{minipage}[b]{\linewidth}\raggedright
\textbf{IQR (Q25--Q75)}
\end{minipage} & \begin{minipage}[b]{\linewidth}\raggedright
\textbf{Frac. low (ρ\textless0.50)}
\end{minipage} & \begin{minipage}[b]{\linewidth}\raggedright
\textbf{Median Top‑3 overlap}
\end{minipage} \\
\midrule
\endfirsthead
\toprule
\begin{minipage}[b]{\linewidth}\raggedright
\textbf{Assay}
\end{minipage} & \begin{minipage}[b]{\linewidth}\raggedright
\textbf{n}
\end{minipage} & \begin{minipage}[b]{\linewidth}\raggedright
\textbf{Median ρ}
\end{minipage} & \begin{minipage}[b]{\linewidth}\raggedright
\textbf{IQR (Q25--Q75)}
\end{minipage} & \begin{minipage}[b]{\linewidth}\raggedright
\textbf{Frac. low (ρ\textless0.50)}
\end{minipage} & \begin{minipage}[b]{\linewidth}\raggedright
\textbf{Median Top‑3 overlap}
\end{minipage} \\
\midrule
\endhead
\textbf{Overall} & \textbf{62} & \textbf{0.90} & \textbf{0.61--1.00} &
\textbf{0.19} & \textbf{1.00} \\
NR‑AR & 4 & 1.00 & 0.94--1.00 & 0.00 & 1.00 \\
NR‑AR‑LBD & 6 & 0.80 & 0.65--0.96 & 0.17 & 0.50 \\
NR‑AhR & 5 & 0.80 & 0.79--1.00 & 0.20 & 1.00 \\
NR‑Aromatase & 6 & 0.75 & 0.45--0.90 & 0.33 & 0.75 \\
NR‑ER & 5 & 0.99 & 0.79--1.00 & 0.00 & 1.00 \\
NR‑ER‑LBD & 5 & −0.23 & −0.64--0.15 & 0.80 & 0.20 \\
NR‑PPAR‑γ & 5 & 0.80 & 0.40--0.94 & 0.40 & 1.00 \\
SR‑ARE & 6 & 1.00 & 0.97--1.00 & 0.00 & 1.00 \\
SR‑ATAD5 & 4 & 0.85 & 0.73--0.92 & 0.00 & 1.00 \\
SR‑HSE & 6 & 1.00 & 0.92--1.00 & 0.17 & 1.00 \\
SR‑MMP & 5 & 0.79 & 0.61--1.00 & 0.20 & 0.50 \\
SR‑p53 & 5 & 1.00 & 1.00--1.00 & 0.00 & 1.00 \\
\bottomrule
\end{longtable}\par}

\clearpage
\section{Supporting Information References}

Assran, M., Duval, Q., Misra, I., Bojanowski, P., Vincent, P., Rabbat,
M., \& Ballas, N. (2023). Self-Supervised Learning from Images with a
Joint Embedding Predictive Architecture IEEE/CVF Conference on Computer
Vision and Pattern Recognition (CVPR), Vancouver,BC,Canada.
https://ieeexplore.ieee.org/document/10205476

Bemis, G. W., and Murcko, M. A. 1996. The Properties of Known Drugs. 1.
Molecular Frameworks. Journal of Medicinal Chemistry, 39(15), 2887-2893.
https://doi.org/10.1021/jm9602928

Chen, M., Radford, A., Child, R., Wu, J., Jun, H., Luan, D., \&
Sutskever, I. (2020). Generative pretraining from pixels Proceedings of
the 37th International Conference on Machine Learning,

Delaney, J. S. 2004. ESOL: Estimating Aqueous Solubility Directly from
Molecular Structure. Journal of Chemical Information and Computer
Sciences, 44(3), 1000-1005. https://doi.org/10.1021/ci034243x

Deng, J., Yang, Z., Wang, H., Ojima, I., Samaras, D., \& Wang, F.
(2023). A systematic study of key elements underlying molecular property
prediction. Nature Communications, 14(1), 6395.
https://doi.org/10.1038/s41467-023-41948-6

Ellington, C., Li, D., Zou, S., Cole, E., Sun, N., Addagudi, S., \&
Xing, E. (2025). Rapid and Reproducible Multimodal Biological Foundation
Model Development with AIDO.ModelGenerator. Obesity, Fitness \& Wellness
Week, 95. https://doi.org/10.1101/2025.06.30.662437

Gilmer, J., Schoenholz, S., Riley, P., Vinyals, O., \& Dahl, G. (2017).
Neural Message Passing for Quantum Chemistry (Vol. 70). JMLR.org.
https://doi.org/10.4316/AECE.2025.01002

Guo, C., Pleiss, G., Sun, Y., and Weinberger, K. Q. 2017. On Calibration
of Modern Neural Networks. Proceedings of the 34th International
Conference on Machine Learning (ICML). https://arxiv.org/abs/1706.04599

Hu, W., Fey, M., Zitnik, M., Dong, Y., Ren, H., Liu, B., Catasta, M., \&
Leskovec, J. (2020). Open graph benchmark: datasets for machine learning
on graphs Proceedings of the 34th International Conference on Neural
Information Processing Systems, Vancouver, BC, Canada.

Hu, W., Liu, B., Gomes, J., Zitnik, M., Liang, P., Pande, V. S., \&
Leskovec, J. (2020). Strategies for Pre-training Graph Neural Networks
https://openreview.net/forum?id=HJlWWJSFDH

Huang, K., Chandak, P., Wang, Q., Havaldar, S., Vaid, A., Leskovec, J.,
\& Zitnik, M. (2024). A foundation model for clinician-centred drug
repurposing (TxGNN). Nature Medicine, 30, 3601--3613.
https://doi.org/10.1038/s41591-024-03233-x

Li, S., Zhou, J., Xu, T., Dou, D., \& Xiong, H. (2021). GeomGCL:
Geometric Graph Contrastive Learning for Molecular Property Prediction
Thirty-Sixth AAAI Conference on Artificial Intelligence, Virtual.
https://ojs.aaai.org/index.php/AAAI/article/view/20377

Liyaqat, T., Ahmad, T., \& Saxena, C. (2025). Advancements in Molecular
Property Prediction: A Survey of Single and Multimodal Approaches. Arch
Computat Methods Eng. https://doi.org/10.1007/s11831-025-10317-5

Piccoli, F., Vogel, G., \& Weber, J. (2025). Joint Embedding Predictive
Architecture for self-supervised pretraining on polymer molecular
graphs. https://arxiv.org/abs/2506.18194

Sanchez-Lengeling, B., Wei, J., Lee, B., Reif, E., Wang, P. Y., Qian, W.
W., McCloskey, K., Colwell, L., \& Wiltschko, A. (2020). Evaluating
attribution for graph neural networks Proceedings of the 34th
International Conference on Neural Information Processing Systems,
Vancouver, BC, Canada.

Schütt, K., Kindermans, P. J., Sauceda, H., Chmiela, S., Tkatchenko, A.,
\& Müller, K. R. (2017). SchNet: A Continuous-Filter Convolutional
Neural Network for Modeling Quantum Interactions. Proceedings of the
31st International Conference on Neural Information Processing Systems,
Long Beach, California, USA.

Segler, M. H., Kogej, T., Tyrchan, C., \& Waller, M. P. (2018).
Generating Focused Molecule Libraries for Drug Discovery with Recurrent
Neural Networks. ACS Cent. Sci., 4(1), 120--131.
https://doi.org/10.1021/acscentsci.7b00512

Simpson, L., Millar, K., Cheng, A., Lim, C.-C., \& Chew, H. G. (2025).
Graph-Based Integrated Gradients for\&nbsp;Explaining Graph Neural
Networks AI 2025: Advances in Artificial Intelligence: 38th Australasian
Joint Conference on Artificial Intelligence, AI 2025, Canberra, ACT,
Australia, December 1--5, 2025, Proceedings, Part I, Canberra, ACT,
Australia. https://doi.org/10.1007/978-981-95-4969-6\_12

Skenderi, G., Li, H., Tang, J., \& Cristani, M. (2024). Graph-level
Representation Learning with Joint-Embedding Predictive Architectures.
https://openreview.net/forum?id=eddd0YTCiq

Soares, E., Shirasuna, V., Brazil, E., Cerqueira, R., Zubarev, D., \&
Schmidt, K. (2024). A Large Encoder-Decoder Family of Foundation Models
for Chemical Language. https://arxiv.org/abs/2407.20267

Soleimany, A. P., Amini, A., Goldman, S., Rus, D., Bhatia, S. N., \&
Coley, C. W. (2021). Evidential Deep Learning for Guided Molecular
Property Prediction and Discovery. ACS Central Science, 7(8),
1356--1367. https://doi.org/10.1021/acscentsci.1c00546

Sterling, T., and Irwin, J. J. 2015. ZINC 15 - Ligand Discovery for
Everyone. Journal of Chemical Information and Modeling, 55(11),
2324-2337. https://doi.org/10.1021/acs.jcim.5b00559

Stokes, J., Yang, K., Swanson, K., Wengong, J., Cubillos-Ruiz, A., \&
Donghia, N. M. (2020). A Deep Learning Approach to Antibiotic Discovery.
Cell, 180(4), 688--702. https://doi.org/10.1016/j.cell.2020.01.021

Sundararajan, M., Taly, A., \& Yan, Q. (2017). Axiomatic attribution for
deep networks Proceedings of the 34th International Conference on
Machine Learning - Volume 70, Sydney, NSW, Australia.

Wang, Y., Li, Z., \& Barati Farimani, A. (2023). Graph Neural Networks
for Molecules. In C. Qu \& H. Liu (Eds.), Machine Learning in Molecular
Sciences (pp. 21--66). Springer International Publishing.
https://doi.org/10.1007/978-3-031-37196-7\_2

Wu, L., Lin, H., Tan, C., Gao, Z., \& Li, S. Z. (2023). Self-Supervised
Learning on Graphs: Contrastive, Generative, or Predictive. IEEE
Transactions on Knowledge and Data Engineering, 35(4), 4216--4235.
https://doi.org/10.1109/TKDE.2021.3131584

Wu, Z., Jiang, D., Hsieh, C.-Y., Chen, G., Liao, B., Cao, D., \& Hou, T.
(2021). Hyperbolic relational graph convolution networks plus: a simple
but highly efficient QSAR-modeling method. Briefings in Bioinformatics,
22(5). https://doi.org/10.1093/bib/bbab112

Wu, Z., Ramsundar, B., Feinberg, E. N., Gomes, J., Geniesse, C., Pappu,
A. S., \& Pande, V. (2017). MoleculeNet: a benchmark for molecular
machine learning. Chemical Science. https://doi.org/10.1039/C7SC02664A

Xu, L., Xia, L., Pan, S., \& Li, Z. (2024). Triple Generative
Self-Supervised Learning Method for Molecular Property Prediction.
International Journal of Molecular Sciences, 25, 3794.
https://doi.org/10.3390/ijms25073794

Ying, R., Bourgeois, D., You, J., Zitnik, M., \& Leskovec, J. (2019).
GNNExplainer: generating explanations for graph neural networks. In
Proceedings of the 33rd International Conference on Neural Information
Processing Systems (pp. Article 829). Curran Associates Inc.

Yuan, Q., Szczypiński, F. T., \& Jelfs, K. E. (2022). Explainable graph
neural networks for organic cages doi: 10.1039/D1DD00039J. Digital
Discovery, 1(2), 127--138. https://doi.org/10.1039/D1DD00039J